\documentclass{article} % For LaTeX2e
\usepackage[preprint]{colm2026_conference}

\usepackage{microtype}
\usepackage{graphicx}
\usepackage{subcaption}
\usepackage{booktabs} % for professional tables
\usepackage{wrapfig}
\usepackage{tcolorbox}
\tcbuselibrary{breakable,skins}
\usepackage{hyperref}
\usepackage{url}
\usepackage{rotating}
\usepackage{diagbox}
\usepackage{amssymb}
\usepackage{multirow}
\usepackage{amsmath}
\usepackage{siunitx}
\usepackage[dvipsnames]{xcolor}
\usepackage{tabularx}

\usepackage{soul}
\usepackage{xcolor}

\definecolor{coco1}{HTML}{D9E4EC}
\definecolor{coco2}{HTML}{B7CFDC}
\definecolor{coco3}{HTML}{6AABD2}
\definecolor{coco4}{HTML}{385E72}
\hypersetup{
    colorlinks=true,
    linkcolor=coco3,
    filecolor=coco4,      
    urlcolor=coco3,
    citecolor=coco3,
}

\usepackage{soul}
\usepackage{xcolor}

\usepackage{xcolor}

\definecolor{tracetext}{HTML}{333333}

\newcommand{\trace}[1]{%
  {\color{tracetext}\small\textit{#1}}%
}

\newcommand{\model}{LongNAP}
\newcommand{\task}{next action prediction}

\newcommand{\gain}[1]{%
  \makebox[3em][r]{\footnotesize\textcolor{ForestGreen}{#1}}\;
}

\def\Snospace~{\S{}}

\usepackage{lineno}

\title{Learning Next Action Predictors\\from Human-Computer Interaction}

\author{Omar Shaikh$^{1}$
  \And
  Valentin Teutschbein$^{2}$\thanks{Equal contribution. Correspondence to \texttt{oshaikh@stanford.edu}.}
  \And
  Kanishk Gandhi$^{1}$\footnotemark[1]
  \AND
  Yikun Chi$^{1}$
  \And
  Nick Haber$^{1}$
  \And
  Thomas Robinson$^{1}$
  \And
  Nilam Ram$^{1}$
  \AND
  Byron Reeves$^{1}$
  \And
  Sherry Yang$^{1,3}$
  \And
  Michael S. Bernstein$^{1}$
  \And
  Diyi Yang$^{1}$
  \AND
  {\normalfont $^{1}$Stanford University \quad $^{2}$Hasso Plattner Institute \quad $^{3}$New York University} 
}

\begin{document}

\ifcolmsubmission
\linenumbers
\fi

\maketitle

\begin{abstract}
Truly proactive AI systems must anticipate what we will do next. This foresight demands far richer information than the sparse signals we type into our prompts --- it demands reasoning over the entire context of what we see and do. We formalize this task as next action prediction (NAP): given a sequence of a user's multimodal interactions with a computer (screenshots, clicks, sensor data), predict that user's next action. Progress on this task requires both new data and modeling approaches. To scale data collection, we annotate longitudinal, naturalistic computer use with vision-language models. We release an open-source pipeline for performing this labeling on private infrastructure, and label over 360K actions across one month of continuous phone usage from 20 users, amounting to 1,800 hours of screen time. We then introduce \model{}, a user model that combines parametric and in-context learning to reason over long interaction histories. \model{} is trained via policy gradient methods to generate user-specific reasoning traces given some context; retrieve relevant traces from a library of past traces; and then apply retrieved traces in-context to predict future actions. Using an LLM-as-judge evaluation metric (0-1 similarity to ground truth), \model{} significantly outperforms supervised finetuning and prompted baselines on held-out data (by 79\% and 39\% respectively). Additionally, \model{} generalizes to held out users when trained across individuals. The space of next actions a user might take at any moment is unbounded, spanning thousands of possible outcomes. Despite this, 17.1\% of \model{}'s predicted trajectories are well-aligned with what a user does next (LLM-judge score $\geq$ 0.5). This rises to 26\% when we filter to highly confident predictions. In sum, we argue that learning from the full context of user behavior to anticipate user needs is now a viable task with substantial opportunity.
\end{abstract}

\section{Introduction}

\begin{figure}
    \centering
    \includegraphics[width=\linewidth]{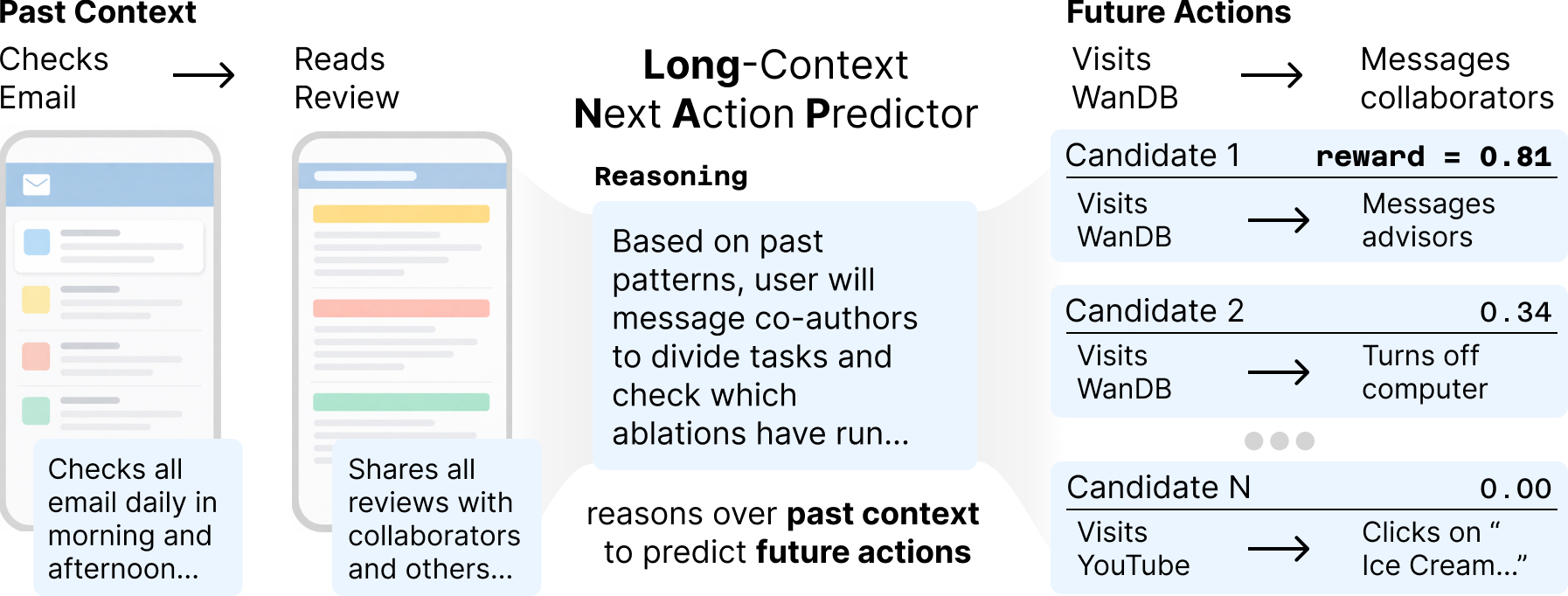}
    \caption{\textbf{Long-Context Next Action Predictors (\model{}s) draw from the entirety of a user's multimodal context (e.g. screenshots)---retrieving over an unbounded history---to predict what they will do next.} We train \model{} end-to-end on data from 20 users over a month, spanning 1.9M screenshots or 1,800 hours of screen on-time.  Predictions are rewarded based on LLM-judged similarity to a set of ground truth future actions.}
    \label{fig:teaser}
\end{figure}

Language models today are hopelessly restricted to seeing us through a narrow keyhole. They see our prompts to follow instructions~\citep{ouyang2022training}, and they construct memories to make sense of these instructions~\citep{packer2023memgpt}. But the models know nothing of what brought us to them in the first place, and they know even less about us in general. Truly context-aware AIs should instead understand us deeply. What problems are we trying to solve? What constraints are we operating under? How do we act in the world, and how can our models best support us in those actions?

To achieve this goal, we need models that learn about us from our general interaction with computers---predicting what we are likely to need or do next---to enable proactive support. These models should reason over a broad history of our past interactions with our devices (not just our past prompts!) in order to forecast what we need help with \emph{before we ask.} 
Success at this task requires training on rich, longitudinal behavioral data, enough to understand our patterns accurately and predict what we will do next. Much of this can run on infrastructure managed or owned by the user, preserving privacy.\footnote{To enable running our work on local infrastructure, code and artifacts for this paper are available at  \url{https://generalusermodels.github.io/nap}.}

We formalize this goal as a concrete prediction task we call \textbf{next action prediction} (NAP): given a sequence of the user's multimodal interactions with a computer (screenshots, keystrokes, clicks), predict that particular user's next action. For example (\autoref{fig:teaser}), consider a researcher who receives a notification, checks their email, then reads a set of paper reviews. A good next action predictor should be able to reason over this sequence, and over what it knows about the user's past habits, to predict that the user will: look through outstanding experiments on their experiment tracking log (e.g., Weights and Biases), then message their helpful coauthors\footnote{We extract this trajectory from the first author's user model, who depends heavily on their helpful coauthors for feedback.} on Slack to divide up revisions.

In this paper, we make progress on two fronts for next action prediction:

\begin{itemize}
    \item \textbf{Data.} How do we collect and annotate the right kind of data for \task{} at scale?
    \item \textbf{Models.} How do we train specialized models for contextual next action prediction that reason effectively over our long-context, multimodal interactions?
\end{itemize} 

Collecting the right data is a prerequisite for progress on this task. Learning accurate models will require large, naturalistic datasets of low-level behavior traces. However, asking users to annotate everything they do is both impractical and expensive. We address this through passive supervision: rather than instructing users to complete specific tasks, we simply observe what they naturally do on their devices, and annotate traces post-hoc using a vision language model (VLM) pipeline. This approach lets us see not what people say they do, but what they actually do. We annotate a dataset of month-long phone use from 20 users, yielding over 360K actions over 1.8K hours of screen on-time. We release this passive data collection pipeline, \textbf{NAPsack}, as an open source package for users to install for themselves.

With a dataset in hand, how should we train models that reason effectively over these long, multimodal interaction histories? A natural approach would be to learn user patterns directly in model weights through finetuning,  building on LLMs that already exhibit social reasoning capabilities~\citep{gandhi2023understanding, ziems2024can}. However, parametric models struggle with latent learning: the ability to acquire and retain information that has no immediate relevance to the current task, but that can be retrieved and applied when it becomes useful for future tasks~\citep{lampinen2025latent}. Patterns in model weights often fail to transfer flexibly to new situations, even when models readily use the same information in-context \citep{chan2022transformers}. Weight updates also require substantially more data than in-context learning to encode new patterns, limiting rapid adaptation to evolving user behavior~\citep{brown2020language, bertsch2025context}. Placing all of a user's interaction history into the context window is also impractical. Context lengths are bounded, and indiscriminately including everything introduces noise that can degrade performance~\citep{liu2024lost}.

\begin{wrapfigure}[22]{r}{0.5\columnwidth}
  \centering
\includegraphics[width=0.48\columnwidth]{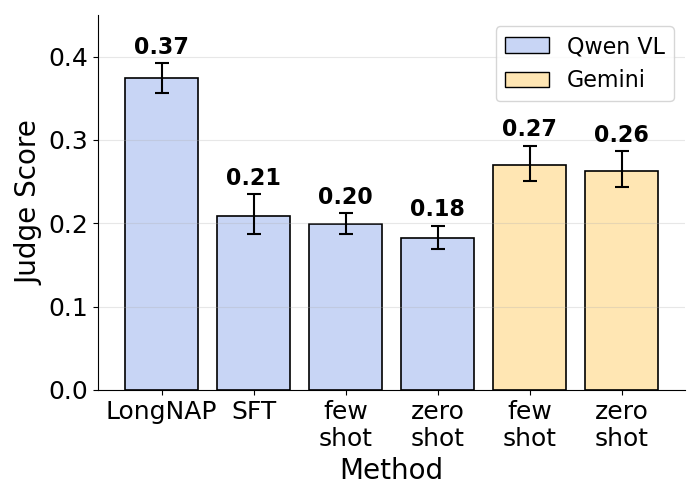}
  \caption{\textbf{\model{} significantly outperforms all baselines by at least 39.4\% relative to the strongest baseline}. We evaluate with LLM-judge, which outputs similarity to ground truth future actions (0-1 score). Performance is averaged across 20 models trained on individual users.}
  \label{fig:llm_judge_bar}
\end{wrapfigure}

Rather than relying solely on parametric or in-context learning, we train models that \emph{learn} to retrieve relevant past reasoning and observations into context, allowing them to leverage strong in-context learning capabilities during the training process. We instantiate this insight in a two-stage model we call \model{} (\textbf{Long}-context \textbf{N}ext \textbf{A}ction \textbf{P}redictor),  trained end-to-end via policy gradient algorithms. In the first phase, \model{} \textbf{reasons to retrieve}: the model reasons about what the user is currently doing, then uses that reasoning to search a memory of past observations and inferences. For example (see \autoref{fig:teaser}), seeing that the user just opened difficult paper reviews, \model{} might retrieve a reasoning trace that it previously generated, noting that the user will message coauthors to divide work. In the second phase, LongNAP \textbf{reasons to predict}: the model integrates retrieved traces to refine its reasoning and predict future actions. Traces that lead to good predictions are saved back into memory, so the library improves over time. To score predictions, we introduce a \textbf{temporal reward}: since we can simply wait and see what the user actually does, we use an LLM-as-a-judge to measure semantic similarity between predicted and actual future actions. This lets us optimize both stages end-to-end through policy optimization.

In our evaluations, we show that \model{}s successfully predict future actions when trained on data from a single user, significantly outperforming supervised finetuning (by 79\%) and prompted baselines (by 39\%). In addition, we show that \model{}s can generalize to entirely \emph{new} users when trained on multiple users, again outperforming baselines (by 13\% over our best baseline---a few-shot prompted, closed-source model). The space of next actions a user might take at any moment is unbounded, spanning thousands of possible outcomes. Despite this, 17.1\% of LongNAP’s predicted trajectories are well-aligned with what a user actually does next (LLM-judge score $\geq$ 0.5 on a 0–1 scale). This rises to 26\% when we filter to highly-confident predictions. 

In sum, we contribute a Long-context Next Action Predictor (\model{}): a model that retrieves and reasons over rich, multimodal interaction histories to predict what a user will do next. To collect data for \model{}, we contribute NAPsack: a pipeline that collects and annotates naturalistic behavior traces passively with VLMs, demonstrating that labeled interaction data can be obtained without any active user effort. \model{}s trained on this data show strong single-user and moderate cross-user generalization. Finally, we discuss applications of \model{}, along with privacy considerations and implications of deploying personalized predictive models on user devices.

\section{Next Action Prediction}

In this section, we formalize the \task{} task and walk through a concrete example. NAP requires operating over a temporal stream of user interaction events
$\mathcal{E} = \{e_1, e_2, \ldots, e_T\}$, where each event
$e_t = (a_t, I_t)$ consists of an action $a_t$ and optional visual observations $I_t$. Here, actions are at the granularity of tasks described in natural language that could be delegated to a computer-use agent \citep{anthropic2024computeruse, xie2024osworld, wang2025opencua}. Below is an example of what $\mathcal{E}$ might look like for a user who needs a NAP:
\[
\mathcal{E} = \left\{\;
\begin{aligned}
& e_1 = (\texttt{snoozes alarm}, \,\texttt{img\_1.png})_{07{:}00}, \\
& e_2 = (\texttt{snoozes alarm}, \,\texttt{img\_2.png})_{07{:}01}, \\
& e_3 = (\texttt{snoozes alarm}, \,\texttt{img\_3.png})_{07{:}02}, \\
& \;\;\vdots
\end{aligned}
\;\right\}
\] 
Given a query time $t$ and a context window containing $k$
recent events $\mathcal{E}_{t-k:t} = \{e_{t-k}, \ldots, e_t\}$,
the goal is to predict future events
$\hat{\mathcal{E}}_{t+1:t+h} = \{\hat e_{t+1}, \ldots, \hat e_{t+h}\}$ that may occur over some horizon $h$, where $h$ and $k$ are parameters set by the user. $\mathcal{E}$ is completely unstructured interaction data---the event stream consists of an arbitrary collection of natural language and images.

We can model this process using a vision-language model (VLM) policy $\pi_\theta$
that generates future event trajectories
$\hat{\mathcal{E}}_{t+1:t+h}$ given recent context $\mathcal{E}_{t-k:t}$. A model trained for contextual behavior prediction would sample from the distribution
$
\hat{\mathcal{E}}_{t+1:t+h}
\sim
p^{\pi}_\theta(\cdot \mid \mathcal{E}_{t-k:t})
$, where $p^{\pi}_\theta(\mathcal{E}_{t-k:t})$ gives the likelihood of a trajectory $\hat{\mathcal{E}}_{t+1:t+h}$ under $\pi$. For example, conditioned on repeated alarm-snoozing behavior, a $p^{\pi}_\theta(\mathcal{E}_{t-k:t})$ might predict the following future trajectory:
\[
\begin{aligned}
\left\{
\begin{aligned}
&(\texttt{snoozes alarm}),\\
&(\texttt{dismisses alarm}),\\
&(\texttt{opens phone})
\end{aligned}
\right\}
\;\sim\;
p^{\pi}_\theta\!\left(
\cdot
\;\middle|\;
\left\{
\begin{aligned}
&(\texttt{snoozes alarm}, \,\texttt{img\_1.png}),\\
&(\texttt{snoozes alarm}, \,\texttt{img\_2.png}),\\
&(\texttt{snoozes alarm}, \,\texttt{img\_3.png})
\end{aligned}
\right\}
\right)
\end{aligned}
\]
Two challenges arise from this problem statement. The first concerns data: how do we scalably collect and annotate a large sample of action data $\mathcal{E}$ across many users? Once we have enough data, we must also effectively train $p^{\pi}_\theta(\mathcal{E}_{t-k:t})$. We cover both challenges in (\autoref{sec:data_collection}; data) and (\autoref{sec:model_outline}; model) respectively.

\section{Labeling Interaction Data At Scale}
\label{sec:data_collection}

Success on \task{} requires large-scale, action-labeled data derived from real computer usage. Such data is scarce due to a high collection cost and practical constraints. It is also impractical to ask individual users to manually segment and annotate everything. 

We draw inspiration from  systems that leverage existing data sources, such as tutorial videos for training computer use agents \citep{baker2022video, wang2025opencua, lu2025videoagenttrek} or passive trajectory data for learning world models in robotics \citep{yang2023learning}. These approaches often rely on curated datasets, synthetic environments, or narrow task distributions. In our setting, we instead want to collect naturalistic, longitudinal, and open-ended computer use that reflects a particular user's behavior. To address this challenge, we introduce \emph{NAPsack}: a passive tool for labeling interaction data from a user at scale.

\subsection{Building NAPsack}
\label{sec:validating_napsack}

NAPsack first continuously records screenshots from computer use; this can include I/O events like mouse clicks, mouse movements, scroll events, and keyboard inputs. I/O events are first grouped into \emph{bursts} of adjacent interactions of the same type (e.g. if a user clicks twice within $\epsilon$ time of each click, it is grouped; see \autoref{appendix:napsack_bursts} for more details). For each burst, NAPsack collects visual context in the form of screenshots of the currently active display before and after the interaction. This serves as a \textbf{compression} strategy; screenshots are only stored when the user actively interacts with the system.

Because bursts corresponding to different event types may temporally overlap, all input events and screenshots are merged into a single, time-ordered sequence.  This sequence is first \textbf{split} into 60-frame chunks, which are provided as input to a VLM. We split to keep context lengths short, since VLMs often forget details as context length increases~\citep{chandrasegaran2024hourvideo}. The VLM is tasked with aggregating one or more consecutive screenshots, optionally annotated with low-level input events that follow (\textbf{IO}, such as keypresses and mouse movements), into higher-level user actions and generating a natural-language action caption. 

Each resulting data sample consists of a screenshot, a generated action description (e.g., \trace{Clicked the `Downloads' folder in the sidebar.}), and the associated input events recorded after that screenshot and before the next one. We intentionally adopt this level of granularity to describe a user's actions because it is better suited for downstream training and directly compatible with prior work on computer-use agents~\citep{wang2025opencua}. To ensure consistent caption granularity and style, we few-shot prompt the VLM (see \autoref{appendix:pack_prompts}).\footnote{One can change the underlying prompts to target higher levels of abstraction in the labeled actions, but there is a tradeoff: higher-level labels require more longitudinal screenshot data.}

\begin{wrapfigure}[33]{r}{0.5\columnwidth}
  \vspace{-1.1em}
  \centering
  \includegraphics[width=0.48\columnwidth]{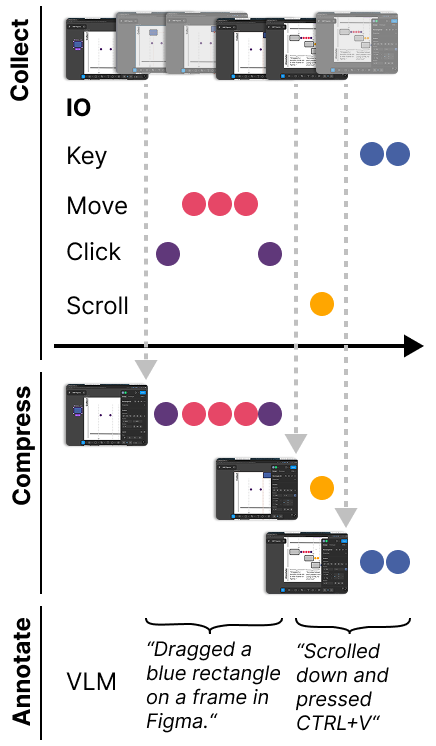}
  \caption{\textbf{NAPsack enables passive collection of human-computer interaction data}. It ingests screenshots and input events, compresses them to retain only meaningful frames, and annotates with action descriptions.}
  \label{fig:pack_schema}
\end{wrapfigure}

\subsection{Evaluating Ground-Truth Similarity with an LLM Judge}
\label{p:pack-llm-eval}

We evaluate NAPsack by comparing generated action captions against ground-truth annotations, testing different data collection and preprocessing strategies for quality and storage efficiency. To benchmark performance, three 10-minute personal computer usage sessions were recorded by an author. Then, two authors produced a total of 354 human-annotated \textbf{ground-truth} action descriptions across the raw screen recordings and input event logs (details in \autoref{appendix:pack_gt})

\paragraph{Conditions} From these recordings, four captioning variants were constructed: (1) \textbf{naively} captioning the full 10-minute inputs with a single prompt, (2) \textbf{splitting} the raw input into 60 frame segments before captioning, (3) applying NAPsack’s event based heuristic for data \textbf{compression} with temporal splitting, and (4) additionally conditioning caption generation on captured input events (\textbf{IO}). 

\paragraph{Similarity with an LLM Judge} For each chunk, we used Gemini~3.0~Flash as a judge, producing a continuous similarity score from 0 (no match) to 1 (perfect match) between ground-truth and a candidate trajectory. Let $\mathcal{H}^*_{t+1:t+h} = \{e^*_{t+1}, \ldots, e^*_{t+h}\}$ denote the ground-truth events that actually occurred after time $t$. Given a candidate set of labels
$\hat{\mathcal{E}}_{t+1:t+h} = \{\hat e_{t+1}, \ldots, \hat e_{t+h}\}$, we can use an LLM to compute a similarity $\mathrm{sim}\!\left(\hat{\mathcal{E}}_{t+1:t+h}, \mathcal{E}^*_{t+1:t+h}\right)$ that measures how well the ground-truth events align with labels produced by each of our baselines. The judge model is prompted to holistically assess similarity to the reference actions. 

Qualitatively, we observe that LLM judges generate more useful scores compared to embedding based or lexical metrics~\citep{zheng2023judging}. To get a sense for judge scores, we highlight examples of similarity scores  in \autoref{tab:judge-examples-horizontal} across various ground-truth/candidate pairs. The exact judge prompt along with additional scored pairs are provided in \autoref{appendix:pack_prompts}.

\begin{table*}[t]
  \centering
  \small
  \newcolumntype{L}{>{\raggedright\arraybackslash}X}
\begin{tabularx}{\textwidth}{@{}c L L L L@{}}
  \toprule
  & \textbf{Ground-Truth} &
  \textbf{Predicted} &
  \textbf{Predicted} &
  \textbf{Predicted} \\
  & \textit{Score = 1.0} &
  \textit{Score = 0.72} &
  \textit{Score = 0.52} &
  \textit{Score = 0.15} \\
  \midrule
  1 & Switched to ``Australia vs Zimbabwe'' tab, entered fullscreen &
      Switched to ``Australia vs Zimbabwe'' tab, entered fullscreen &
      Switched to ``Australia vs Zimbabwe'' tab, clicked player &
      \textcolor{red}{Clicked on YouTube home page, browsed recommended videos} \\ \noalign{\vspace{6pt}}
  2 & Clicked thumbnail in ``Key Moments'' section &
      Clicked on video player timeline to skip forward &
      Pressed ``f'' to enter fullscreen on cricket video &
      Clicked video titled \textcolor{red}{``Best Cricket Catches of 2025''} \\ \noalign{\vspace{6pt}}
  3 & Pressed ``esc'' to exit fullscreen &
      Pressed ``esc'' to exit fullscreen &
      Pressed ``esc'' to exit fullscreen &
      Pressed ``esc'' to exit fullscreen \\ \noalign{\vspace{6pt}}
  4 & Switched to W\&B workspace, clicked ``Runs'' icon &
      Switched to W\&B workspace, \textcolor{red}{scrolled through dashboard} &
      Opened new tab, typed ``wandb.ai'' into address bar &
      \textcolor{red}{Typed ``australia cricket schedule'' into Google search bar} \\ \noalign{\vspace{6pt}}
  5 & Switched to ``Australia vs Zimbabwe,'' entered fullscreen &
      Switched to ``Australia vs Zimbabwe'' tab, entered fullscreen &
      Scrolled through W\&B charts for training metrics &
      \textcolor{red}{Clicked ESPN Cricinfo link in search results} \\ \noalign{\vspace{6pt}}
  6 & Pressed ``esc'' to exit fullscreen &
      Pressed ``esc'' to exit fullscreen &
      \textcolor{red}{Clicked ``Table'' view in W\&B dashboard} &
      \textcolor{red}{Scrolled through upcoming matches on ESPN Cricinfo} \\ \noalign{\vspace{6pt}}
  7 & Switched to W\&B tab, scrolled through charts &
      Switched to W\&B tab, \textcolor{red}{clicked} on a chart &
      \textcolor{red}{Switched to YouTube, clicked recommended video} &
      \textcolor{red}{Clicked back button in Chrome to return to search results} \\ \noalign{\vspace{6pt}}
  8 & Scrolled through W\&B dashboard &
      \textcolor{red}{Clicked ``Table'' view in W\&B workspace} &
      \textcolor{red}{Scrolled through comments section on YouTube} &
      \textcolor{red}{Switched to Gmail tab and checked for new emails} \\
  \bottomrule
  \end{tabularx}
  \caption{%
    \textbf{Qualitative examples of candidate trajectories $\hat{\mathcal{E}}$ at decreasing LLM judge similarity scores (left to right),
    all evaluated against the same ground-truth $\mathcal{H}^*$ human annotated trajectory.} For readability, we highlight in \textcolor{red}{red} actions with substantially wrong intent.
  }
\label{tab:judge-examples-horizontal}
\end{table*}
 
\paragraph{LLM Judge Results} For LLM judge results, we first sample a trajectory of 8 ground-truth, human-annotated labels $\mathcal{H}^*$, along with a candidate subsequence that covers the same timespan $\hat{\mathcal{E}}$ from each condition. We generate 45 ground-truth sequences, each with a complementary candidate sequence from each condition. \autoref{tab:pack-eval} reports mean LLM-as-a-judge scores comparing each candidate to the ground truth, together with the corresponding data storage requirements.\footnote{Filesizes are computed by concatenating final selected frames into an .mp4 video.}

Our judge shows that splitting input (\texttt{+~split}) substantially improves caption quality compared to providing the entire session at once (\texttt{naive}; $0.48 \to 0.57$). Using only frames where a user interacts with their computer (\texttt{+~compress}) achieves comparable caption quality while reducing storage by approximately $70\%$ ($\SI{295}{\mega\byte} \to \SI{76}{\mega\byte}$). Finally, conditioning on input events further improves scores (\texttt{IO}; $0.60 \to 0.70$); I/O data may provide complementary supervision beyond visual context alone.

\paragraph{Human Validation}
\label{sec:judge_val}
To validate our LLM judge, two authors independently labeled pairwise preferences across conditions. For each pair, annotators were shown a ground-truth sequence alongside one sample from each condition and asked to select the better match (ties counted as 0.5 wins for each). Altogether, this results in 240 total comparisons across both annotators. We report aggregate win rates averaged across both annotators (\autoref{tab:pack-eval}).

Human win-rates validate our LLM judge evaluation. Splitting inputs provides a significant improvement over naively providing an entire video session as input ($19.2\% \to 45.4\%$); compression does not significantly affect quality ($45.4\% \to 49.6\%$); and adding IO input significantly improves quality ($49.6\% \to 85.8\%$). We attribute these gains to the ability to capture short, keyboard-heavy interactions (single terminal commands or window switching) when I/O is available, a detail overlooked in purely frame-based variants. Together, our evaluation justifies NAPsack's use for future scaling of accurate, action-labeled trajectories.

\begin{table}[]
    \centering
    \begin{tabular}{lrrrr}
        \toprule
        Method & Judge Score ($\uparrow$) & Human WR \% ($\uparrow$) & Size ($\downarrow$) \\
        \midrule
        naive & 0.48 $\pm$ 0.03 & 19.2 $\pm$ 6.7 & \SI{295}{\mega\byte} \\
        \midrule
        + split & 0.57 $\pm$ 0.03 & 45.4 $\pm$ 8.7 & \SI{295}{\mega\byte} \\
        + split + compress & 0.60 $\pm$ 0.03 & 49.6 $\pm$ 8.9 & \SI{76}{\mega\byte}\\
        + split + compress + IO & \textbf{0.70 $\pm$ 0.03} & \textbf{85.8 $\pm$ 6.0} & \SI{76}{\mega\byte}\\
        \bottomrule
    \end{tabular}
    \caption{\textbf{NAPsack reduces the amount of data we have to save for effective captioning by 75\% without compromising quality}. Event-driven compression (where frames are saved only when a user interacts with their computer) yields improved efficiency; and I/O signals further improve performance. Performance is measured across both LLM-judge scores ([0, 1]) and human eval (win rate). 95\% conf. intervals are computed through bootstrapping.}
    \label{tab:pack-eval}
\end{table}

\subsection{Annotating a Dataset with NAPsack}
\label{sec:screenomics_annotate}
We now turn to annotating a large-scale dataset for training models and experimentation.

\paragraph{Screenomics} Collecting a large scale dataset of real-world computer interaction is challenging. As a starting point, we draw on Screenomics~\citep{reeves2021screenomics}, and a study from the Human Screenone Project~\citep{reeves2020time}, a repository of $\approx$ 170M continuous timelapse screenshots collected from 257 adult users' mobile phone activity over time. Given the highly sensitive nature of the data, collection and annotation of the Screenomics data was first approved by the Stanford Institutional Review Board. All modeling was done on secure servers approved for processing personal data at the first author's institution.

\paragraph{Subsampling Screenomics} We begin by sampling a set of 20 users who used their devices for at least an hour every day for a month. Demographic details on the 20 users are in \autoref{appendix:screenomics_users}. Our final sampled time window occurs between March 16, 2021 and April 12, 2021, and consists of 1.9M screenshots, covering 1,837 hours of total screen on-time.
  
\paragraph{Annotating with NAPsack} We feed each user's screenshot data into NAPsack, yielding a total of 360K event descriptions. Screenomics does not capture IO data, so we omit this from the input to NAPsack, compressing instead with a difference in image hash (see \autoref{appendix:screenomics_users}). On average, each event description $\mathcal{E}_i$ covers $\approx$ 15 seconds of time, yielding a total of 359,219 actions. Events describe a diverse set of activities, from gaming and shopping, to banking, messaging, and social media browsing. Across 20 users over 28 days, average daily screen time was 4h 32m, ranging from 2h 17m (lightest) to 6h 52m (heaviest). While we validate our captioning processes, labels from NAPsack are model-generated, and can still be noisy. Alignment with human ground-truth is not perfect. This may introduce errors later in training; we revisit this in our limitations section (\autoref{sec:disc_limit}).

\section{Long-Context Next Action Predictors (\model{})}
\label{sec:model_outline}

To predict what a user might do next, we must be able to reason effectively over the entirety of their context---retrieving and re-using specific details or insights across potentially infinite observations. This is challenging with just weight-based learning, as a model must be able to use new information immediately: if a user checks their calendar and sees a 2 P.M. meeting, that detail should inform predictions right away, not after further gradient updates. Our model must learn from a single observation, and many relevant details (a new appointment, a message from a collaborator) appear only once and are never repeated in the training data~\citep{chan2022transformers}.

We instead exploit the ability of LLMs to quickly adapt via in-context learning~\citep{brown2020language}. While it would be ideal for the context window to be unbounded, we are constrained by practical context limitations of LLMs. Thus, we design a learning architecture for \model{}s and train them with a two phase generation process. They \emph{reason to retrieve} relevant past context (old observations and reasoning traces); then \emph{reason to predict} the final set of next actions. 

\begin{figure}[tbp]
    \centering
    \includegraphics[width=\linewidth]{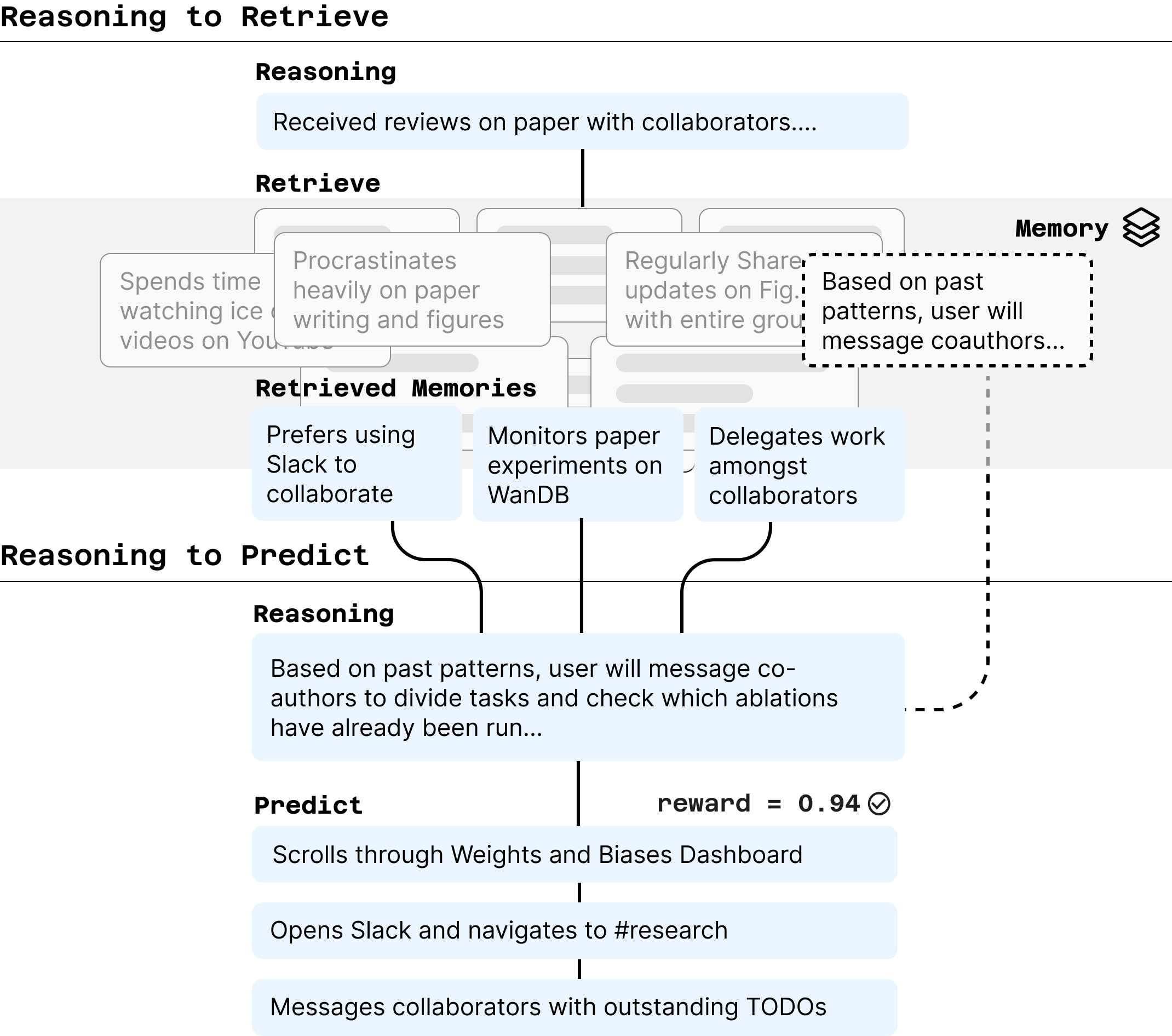}
    \caption{\textbf{Predictions from \model{} are generated in a two-phase process.} In the first phase, \model{} \textbf{Reasons to Retrieve}: conditioned on what the user sees right now (e.g. a set of paper reviews), \model{} generates a reasoning trace and uses it to retrieve past traces. Using retrieved traces, \model{} \textbf{Reasons to Predict}: generating a final reasoning trace, adding it back to memory, and then predicting the next steps a user might take. The predicted trajectory is compared against a ground truth (with an LLM-as-a-judge similarity score), and then \model{} is optimized with GRPO~\citep{shao2024deepseekmath}.}
    \vspace{-2mm}
    \label{fig:longnap}
\end{figure}
\subsection{Reasoning to Retrieve and Predict}

Implementing this generation requires a few prerequisites. First, we need a VLM policy $\pi$---we instantiate this VLM using Qwen-2.5-VL-7B~\citep{bai2025qwen2}. \model{} maintains a memory $\mathcal{M}_t$ of past entries available up to time $t$. Each memory entry pairs a set of observations with a \textit{reasoning trace}: a chain-of-thought, $z$, generated by the model during a previous prediction (e.g., \trace{User received paper reviews;} \trace{based on past behavior, they tend to procrastinate on writing but eventually coordinate with coauthors via Slack}; \autoref{fig:teaser}). To search over this memory, we instantiate a lexical retriever $\mathcal{R}$, using BM25~\citep{robertson1995okapi}. Only entries with timestamps $\tau \leq t$ are accessible, preventing access to future information. The policy $\pi$ is then tasked with continuously retrieving from and updating this memory as new events occur and new reasoning traces are generated. Below, we walk through how LongNAP processes the example in \autoref{fig:longnap}.

\paragraph{Phase 1: Reasoning to Retrieve.}
Suppose a user browses aimlessly, receives a notification, checks their email, and reads a set of paper reviews. Given these $k$ recent observations, the model first generates reasoning about what might come next: $z^{\text{retrieve}} \sim p_\theta(\cdot \mid \mathcal{E}_{t-k:t})$ (\autoref{fig:longnap}; top). Reasoning traces from the model speculate on the user's context and stable traits; in our example, the model might reason: \trace{Received reviews on paper with collaborators\ldots user may revise paper after viewing feedback}. This reasoning serves a dual purpose: it makes the model's current thinking explicit, and it provides a semantic query for retrieving relevant history. Using $z^{\text{retrieve}}$ as a query, we retrieve entries $D = \mathcal{R}(z^{\text{retrieve}}, \mathcal{M}_t)$ from memory. Here, the retriever might surface past (abridged) traces such as \trace{Procrastinates heavily on paper writing and figures}, \trace{Prefers using Slack to collaborate}, and \trace{Delegates work amongst collaborators}.

\paragraph{Phase 2: Reasoning to Predict.}
The model then revises its initial prediction by integrating the retrieved context (\autoref{fig:longnap}; bottom): $z^{\text{predict}} \sim p_\theta(\cdot \mid \mathcal{E}_{t-k:t}, z^{\text{retrieve}}, D)$. In Phase~1, the model already speculated about what comes next, but it did so without any historical context from the memory about the user. Now, the retrieved traces about this user's procrastination habits, preference for Slack, and tendency to delegate allow the model to revise. What started as a generic \trace{Received reviews \ldots user may revise paper after viewing feedback} becomes: \trace{Based on past patterns, user will message coauthors to divide tasks, check which experiments have  been run}. Conditioned on this revised reasoning, the model predicts concrete future actions: $\mathcal{E}_{t+1:t+h} \sim \pi_\theta(\cdot \mid \mathcal{E}_{t-k:t}, D, z^{\text{predict}})$, predicting the user will \trace{scroll through Weights \& Biases}, \trace{open Slack} and \trace{navigate to \#research}, and \trace{message collaborators with outstanding TODOs.}

During training, we sample 4 candidate traces for each phase, producing 4 complete retrieve-and-predict rollouts. After prediction, we save the prediction trace with the highest reward back to memory, updating $\mathcal{M}_{t+1} = \mathcal{M}_t \cup \{(\mathcal{E}_{t-k:t}, z^{\text{predict*}}, t)\}$. This ensures the memory accumulates the model's best reasoning over time.

\subsection{Optimizing \model{}s} 
\model{} is trained end-to-end, learning how to generate initial reasoning, what to retrieve from memory, and how to revise its reasoning for accurate predictions. Because generation involves discrete steps (reasoning in language, calling a retriever), we optimize via policy gradients, using GRPO~\citep{shao2024deepseekmath, liu2025understanding} for variance reduction with a group size of 4. We use LoRA~\citep{hu2022lora} due to memory constraints, where RL results generally match full finetuning~\citep{schulman2025lora}. Additional hyperparameter details for both training and calling our retriever are in \autoref{appendix:longnap_hyperparameters}

\paragraph{Temporal Reward Formulation} The temporal structure of our task provides a natural training signal: we can verify rollout quality by comparing predicted actions against observed future behavior. In other words, we can just wait and see if the user does what we predict. Here, we re-use the same validated LLM Judge for NAPsack (in \autoref{sec:judge_val}, Gemini 3.0 Flash as the underlying LLM) and apply this as our reward. To allow the LLM to distinguish between each completion more effectively, we also pass the entire group at once to the model, and prompt the model to assign rewards all at once. 

\paragraph{Training With Memory} There are a handful of complications that come with adding memory. First, we want historical reasoning and observations to accrue over time in memory. Predicting what a user will do next by starting immediately in the middle of the dataset (e.g. after a shuffle) is challenging. We instead train over the data chronologically, and reset memory at the end of each epoch. Second, we \emph{mask} retrieved tokens since they come from the environment, following search engine tool-use~\citep{jin2025search}. Finally, we apply a form of ``dropout'' to our retriever. We randomly drop (10\% of the time), re-order (10\%), or provide no items (10\%) as context. We find that this generally stabilizes training, preventing collapse when memory is reset at the start of an epoch.

\section{Experimental Setup}
With our dataset (\autoref{sec:screenomics_annotate}) and model (\autoref{sec:model_outline}), we turn to evaluating \model{}s in two settings. First, we evaluate if a \model{} trained on a single user generalizes over \emph{time}, predicting what that single user will do in the future. Second, we test if \model{}s trained on many users generalize to entirely new ones. In both settings, we evaluate \model{} against prompted and supervised-finetuned baselines, and show that \model{} significantly outperforms baselines.

\subsection{Experiments and Evaluation Splits}
\label{sec:splits_exps}
\paragraph{Prediction Event Horizon} Before we outline experiments, we fix a few parameters for consistency across experiments. \model{} samples $
\hat{\mathcal{E}}_{t+1:t+h}
\sim
p^{\pi}_\theta(\cdot \mid \mathcal{E}_{t-k:t}).
$, so we need to define both how many events we should predict $h$ \emph{and} how many events should be in the context window $k$. We take a sliding window over all our ground truth events $\mathcal{E}$ to generate this dataset. Both the future action horizon $h$ and past actions $k$ are hyperparameters that can change depending on the prediction task. For now, we set the context window to 16 events, and future prediction to 8 events. We leave exploring different horizons and contexts to future work.

\paragraph{Generalizing Over Time} Here, we aim to understand if \model{} generalizes over time, predicting actions one user might do in the future. This requires training 20 models, one for each participant. We split temporally within participants: the first two weeks of data are for training (9.1K actions on average per user), the third week for validation (4.4K), and the fourth week for test (4.4K). 
  
\paragraph{Generalizing to New Users} In this setup, we aim to see if \model{} can generalize to entirely new users. We train a single model over many users, and then evaluate on new, unseen users. To do this, we split our annotated data of 20 users into 10 randomly-selected users in train, 5 in validation, and 5 in test. While we have a single policy $\pi_{\theta}$, we cannot share memory between users, so we also instantiate a separate memory per-user (e.g. 10 separate memories are maintained during training). During the generation process, a model only indexes into the retriever for the specific user. 

\subsection{Automated Metrics and Human Validation} 
\label{sec:metrics}
To measure the closeness of our predicted events $\hat{\mathcal{E}}_{t+1:t+h}$ to the ground truth $\mathcal{E}^*_{t+1:t+h}$, we employ two metrics. First, we again re-use our \textbf{validated LLM-judge} for comparing similarity between ground truth and predicted trajectories (\autoref{sec:judge_val}). The LLM judge provides us with a more granular sense for low-level correctness. To understand the upper bound in \model{} performance, we also report \textbf{pass@k} performance. The pass@k evaluation involves drawing k samples (temp = 1.0) from each model, and scoring an instance ``correct'' if any of the k samples is deemed close enough to the ground truth. Here, we pick a high cut-off for our LLM judge score (judge $>$ 0.50). 

At a threshold of 0.50, our calibrated judge (\autoref{sec:judge_val}) indicates that the predicted and actual trajectories share the same actions, though some details or ordering may differ. For example, a predicted trajectory of \trace{Opens Chrome, navigates to a Weights \& Biases dashboard, adjusts a chart slider, inspects training metrics} and a ground truth of \trace{Opens Chrome, scrolls through the Weights \& Biases dashboard, clicks into a specific pipeline chart, switches to the terminal} are rated 0.6—the core workflow (examining experiment metrics) overlaps, but specific interactions and subsequent steps differ. In contrast, \trace{Opens YouTube, watches music videos, browses recommended content} and \trace{Opens Chrome, analyzes experiment dashboards, switches to the terminal to debug a pipeline} are rated 0.0, with no meaningful overlap in intent or activity. A cutoff of 0.50 for our judge enables us to roughly identify which trajectories are mostly correct.

We also measure \textbf{model confidence} for a given trajectory. We compute the intra-cluster variance of 20 sampled predictions. First, we embed each sample using a sentence transformer (using the all-MiniLM-L6-v2 model; \citet{wang2020minilm,reimers2019sentence}). We  then compute the average squared Euclidean distance from each embedding to their centroid (similar to \cite{farquhar2024detecting}). Lower variance indicates higher agreement among samples, which we interpret as higher model confidence. We convert these values to per-user percentile ranks to account for individual differences in baseline spread.

Finally, we conduct a small-scale \textbf{human eval} to validate our LLM judge. Two authors independently labeled pairwise preferences across methods, resulting in 300 total comparisons. For each pair, annotators were shown a ground-truth sequence alongside one output from each method and asked to select the better match based on overall quality (ties counted as 0.5 wins for each). Pairs were sampled in a stratified fashion across users to ensure each user is well-represented. We report aggregate win rates averaged across both annotators.

\subsection{Baselines} Our baselines consist of closed and open models, both across prompting  and finetuning methods. For prompting, we have a \textbf{zero-shot} baseline, where we give the model the immediate past actions the user took, and simply prompt it to predict what the user would do next. We additionally implement a basic \textbf{few-shot RAG} baseline, where we use past actions as a query for retrieving. Prompts for the above baselines are in \autoref{appendix:prompt_baselines}. For finetuned baselines, we evaluate \textbf{supervised finetuning}, testing if simply finetuning over the set of actions is helpful. For closed models, we evaluate prompted baselines with Gemini's 3.0 Flash.\footnote{We use Flash since preliminary experiments show marginal improvements over Pro for a fraction of the cost.} For open models, we prompt and finetune Qwen-2.5-VL-7B~\citep{bai2025qwen2}.
\section{Results}
\label{sec:overall_results}

\begin{table*}[t]
\centering
\renewcommand{\arraystretch}{1.05}
\begin{tabular}{l|rr|rrrr}
\toprule
User
& \multicolumn{2}{c|}{Gemini}
& \multicolumn{4}{c}{Qwen-2.5-VL-7B} \\
$u_i$ & Zero-shot & RAG
& Zero-shot & RAG & SFT & LongNAP \\
\midrule

$u_{1}$  & 0.28 & 0.29 & 0.21 & 0.23 & 0.23 & \gain{+0.12}\textbf{0.41} \\
$u_{2}$  & 0.21 & 0.22 & 0.15 & 0.17 & 0.18 & \gain{+0.12}\textbf{0.34} \\
$u_{3}$  & 0.23 & 0.23 & 0.19 & 0.18 & 0.17 & \gain{+0.10}\textbf{0.33} \\
$u_{4}$  & 0.24 & 0.23 & 0.14 & 0.16 & 0.17 & \gain{+0.09}\textbf{0.32} \\
$u_{5}$  & 0.23 & 0.24 & 0.18 & 0.16 & 0.16 & \gain{+0.10}\textbf{0.34} \\
$u_{6}$  & 0.24 & 0.25 & 0.16 & 0.20 & 0.19 & \gain{+0.12}\textbf{0.37} \\
$u_{7}$  & 0.40 & 0.42 & 0.20 & 0.22 & 0.40 & \gain{+0.04}\textbf{0.46} \\
$u_{8}$  & 0.23 & 0.24 & 0.15 & 0.18 & 0.17 & \gain{+0.15}\textbf{0.39} \\
$u_{9}$  & 0.25 & 0.25 & 0.18 & 0.21 & 0.22 & \gain{+0.12}\textbf{0.37} \\
$u_{10}$ & 0.29 & 0.29 & 0.24 & 0.26 & 0.24 & \gain{+0.10}\textbf{0.39} \\
$u_{11}$ & 0.26 & 0.26 & 0.17 & 0.17 & 0.19 & \gain{+0.04}\textbf{0.30} \\
$u_{12}$ & 0.25 & 0.27 & 0.18 & 0.22 & 0.26 & \gain{+0.13}\textbf{0.40} \\
$u_{13}$ & 0.26 & 0.26 & 0.18 & 0.20 & 0.19 & \gain{+0.11}\textbf{0.37} \\
$u_{14}$ & 0.27 & 0.28 & 0.20 & 0.21 & 0.18 & \gain{+0.08}\textbf{0.36} \\
$u_{15}$ & 0.23 & 0.25 & 0.17 & 0.20 & 0.20 & \gain{+0.11}\textbf{0.36} \\
$u_{16}$ & 0.24 & 0.24 & 0.17 & 0.18 & 0.21 & \gain{+0.17}\textbf{0.41} \\
$u_{17}$ & 0.25 & 0.25 & 0.17 & 0.18 & 0.15 & \gain{+0.10}\textbf{0.34} \\
$u_{18}$ & 0.27 & 0.27 & 0.17 & 0.22 & 0.16 & \gain{+0.11}\textbf{0.38} \\
$u_{19}$ & 0.41 & 0.41 & 0.28 & 0.27 & 0.30 & \gain{+0.05}\textbf{0.46} \\
$u_{20}$ & 0.24 & 0.26 & 0.15 & 0.17 & 0.24 & \gain{+0.15}\textbf{0.41} \\

\midrule
$u_{\mu}$ & 0.26 & 0.27 & 0.18 & 0.20 & 0.21 & \gain{+0.11}\textbf{0.38} \\
\bottomrule
\end{tabular}
\caption{\textbf{Training on individual users substantially outperforms baseline methods, improving average performance by 39.4\% (\gain{+0.11}) relative to the strongest baseline (Gemini Few-shot RAG).} We report the similarity score (0-1) to ground truth future actions, as determined by our LLM judge (\autoref{sec:judge_val}). $u_i$ denotes a \model{} instance trained on a single user from our dataset of 20 users, and $u_{\mu}$ represents the mean performance across all 20 individually trained models.}
\label{tab:single_user_results}
\end{table*}

\begin{figure}
    \centering
    \includegraphics[width=\linewidth]{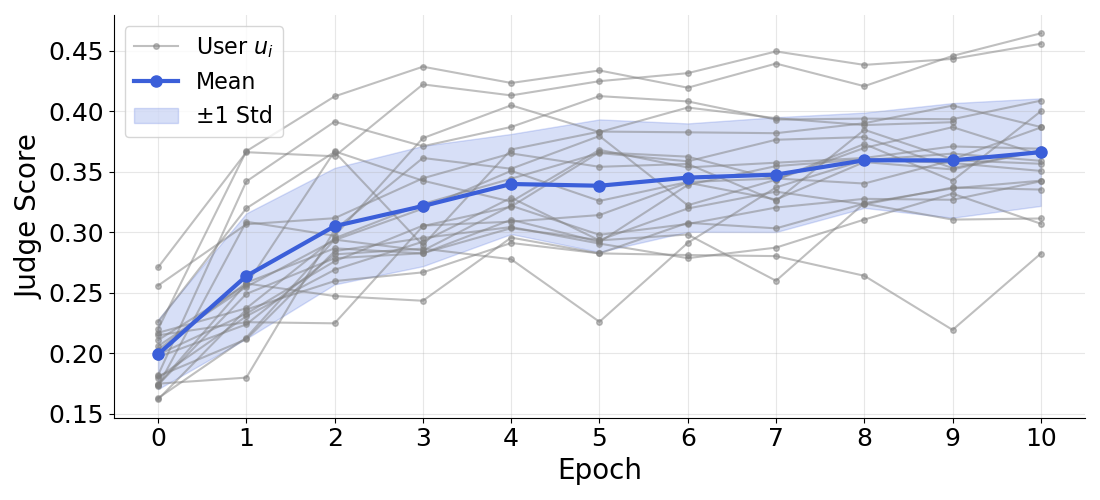}
    \caption{\textbf{Some users are more predictable than others.} When \model{} is trained on a single user (in our generalizing over time experiments, \autoref{sec:splits_exps}), LLM-as-a-judge evals vary substantially from one user's \model{} to another user's. In the above figure, we re-evaluate across checkpoints from training epochs for each user, highlighting variance.}
    \label{fig:val_scores_plot}
\end{figure}

We synthesize takeaways from our evaluation, with main results summarized in  \autoref{tab:single_user_results} and \autoref{tab:many_user_results}.

\begin{table}[t]
\centering
\begin{tabular}{llr}
\toprule
\textbf{Model} & \textbf{Method} & \textbf{Win Rate (\%)} \\
\midrule
Qwen 2.5 VL & \model{} & \textbf{79.0 $\pm$ 6.6} \\
& SFT               & $29.5 \pm 6.9$ \\
& Few-shot RAG               & $45.0 \pm 7.6$ \\
& Zero-shot         & $33.0 \pm 7.2$ \\
\midrule
Gemini & Few-shot RAG & $58.5 \pm 6.8$ \\
 & Zero-shot          & $55.0 \pm 7.7$ \\
\bottomrule
\end{tabular}
\caption{\textbf{Human evaluation shows significantly higher win rates for \model{}.} We show pairwise human evaluation win rates across each method (combined across 2 annotators, 300 total comparisons, 95\% bootstrapped CIs).}
\label{tab:method-win-rates}
\end{table}

\paragraph{\model{} can learn from just a single user, predicting their future interaction} When we train on just a single user, \model{} is able to predict what the user does next with significantly higher accuracy compared to baselines. When we compare against SFT on Qwen-2.5-VL-7B, \model{} achieves \textbf{79\%} higher performance (0.21 $\rightarrow$ 0.38). \model{} also substantially outperforms prompted baselines. \model{} achieves \textbf{106\%} higher performance than zero-shot prompting (0.18 $\rightarrow$ 0.38) and \textbf{88\%} higher performance compared to few-shot prompting (0.20 $\rightarrow$ 0.38). These gains extend to closed-source models: \model{} achieves \textbf{43\%} and \textbf{39\%} higher performance than zero-shot (0.26 $\rightarrow$ 0.38) and few-shot (0.27 $\rightarrow$ 0.38) prompted Gemini 3.0 Flash, respectively.

Our human evaluation further validates these findings (\autoref{tab:method-win-rates}). \model{} achieves a \textbf{79\%} win rate against other methods, substantially outperforming SFT (29.5\%), zero-shot (33\%), and RAG (45\%) baselines on Qwen 2.5 VL. Notably, \model{} also surpasses stronger closed-source baselines, beating both zero-shot (55\%) and RAG-prompted (59\%) Gemini.

\paragraph{\model{} can learn from many users, generalizing to entirely \emph{new} users.} We find that \model{}, when trained on many users at once, generalizes to new users. While the gains are not as large as in the single-user setting, \model{} still substantially outperforms prompted open-source baselines. In particular, \model{} achieves \textbf{66\%} higher performance than zero-shot prompting Qwen-2.5-VL-7B and \textbf{53\%} higher performance than few-shot prompting. Improvements over the closed-source baseline (Gemini 3.0 Flash) are more modest: \model{} achieves \textbf{19\%} and \textbf{13\%} higher performance than zero-shot and few-shot prompting, respectively. These are smaller gains relative to the single-user setting, so we suspect user-specific weights are especially effective for NAP. Scaling users may close this gap; we leave this to future work. 

We also suspect that this variant of \model{} relies more heavily on the \textbf{reasoning to retrieve} process. At inference time, \model{} must learn general strategies for saving and retrieving user-specific inferences, depending less on parametric memorization and more on retrieval and in-context learning. 

\paragraph{LongNAP's most confident predictions are aligned with what users actually do 26\% of the time.} To provide a more interpretable measure of performance, we report pass@k: the probability that at least one of $k$ independent samples from a model exceeds a similarity threshold against the ground truth future trajectory. We selected an LLM-judge threshold of 0.5; trajectories that get this score are often well aligned with the actual intent of the user, but miss details or skip a few actions (see \autoref{sec:metrics} for an example). At this threshold, LongNAP achieves \textbf{17.1\% at pass@1} across users, rising to \textbf{36.3\% at pass@20} (\autoref{fig:pass_at_k}). In addition, we observe that model confidence, measured as intra-cluster variance among the 20 sampled trajectories, is correlated with accuracy (\autoref{fig:empiric_calib}). For prompts in the 90th percentile of confidence (lowest variance), \textbf{pass@1 rises to 25.9\%}.

\begin{table}[]
    \centering
    \begin{tabular}{llr|rrrrr}
        \toprule
        Model & Method & $u_{\mu}$ & $u_{2}$ & $u_{6}$ & $u_{12}$ & $u_{14}$ & $u_{16}$ \\ 
        \midrule
      \multirow{2}{*}{Gemini}
        & Zero-shot  & 0.22 & 0.27 & 0.22 & 0.20 & 0.20 & 0.21 \\
        & Few-shot RAG & 0.23 & 0.26 & 0.23 & \textbf{0.22} & \textbf{0.21} & 0.22 \\
       \midrule
      \multirow{4}{*}{Qwen VL}
        & Zero-shot  & 0.16 & 0.22 & 0.13 & 0.14 & 0.16 & 0.14 \\
        & Few-shot RAG & 0.17 & 0.23 & 0.15 & 0.18 & 0.14 & 0.18 \\
        & SFT & 0.17 & 0.22 & 0.14 & 0.16 & 0.15 & 0.17 \\
        & LongNAP & \textbf{0.26} & \textbf{0.31} & \textbf{0.28} & 0.22 & 0.21 & \textbf{0.26} \\
       \bottomrule
    \end{tabular}
    \caption{\textbf{When trained jointly on many users, \model{} exhibits modest generalization to unseen users, improving average performance by 13.0\% relative to the strongest baseline (again, Gemini Few-Shot RAG).} We report similarity to ground truth future actions, as determined by our LLM judge (\autoref{sec:judge_val}). To test generalization to unseen users, we train randomly on 10 users; validate on 5; and then test on 5 (reported above). $u_{\mu}$ denotes the mean performance across the evaluated users, and $u_i$ represents performance on user $i$.}
  \label{tab:many_user_results}
\end{table}

\paragraph{Predictability across users and generalization.} While \model{} results are on average better than our baselines, there is still substantial variability across users. For some users, relative improvement against the strongest baseline is limited, while for others, improvement is substantial. On $u_{11}$, for example, the strongest baseline is Gemini (few-shot), which achieves a score of 0.26 with the LLM judge; \model{} improves modestly to 0.30 -- an absolute gain of 4 points (\textbf{15\%} relative improvement). In contrast, on $u_{8}$, the strongest baseline is Gemini few-shot (0.24), and \model{} reaches 0.39, corresponding to a 15-point absolute gain (\textbf{63\%} relative improvement). We suspect that some users are inherently more predictable. They repeat similar tasks each day, making them easier to model with finetuning or prompting alone, limiting the additional benefit of RL~\citep{chu2025sft}. 
\section{What Makes \model{} Work?}
\label{sec:ablations}

\begin{table} %[htb!]
  \centering
  \renewcommand{\arraystretch}{1.15}
  \resizebox{\linewidth}{!}{%
\begin{tabular}{l r| rrrrr}
    \toprule
      {Model} & {$u_{\mu}$} & {$u_{2}$} & {$u_{6}$} & {$u_{12}$} & {$u_{14}$} & {$u_{16}$} \\
    \midrule
    \model{}
      & \num{0.38}
      & \num{0.34}
      & \num{0.37}
      & \num{0.40}
      & \num{0.36}
      & \num{0.41} \\
    \midrule
    \multicolumn{7}{l}{\textit{Ablations}} \\
    \midrule
   $\rightarrow$ Remove reasoning
      & {\footnotesize\textcolor{red}{(-0.07)}}\;\num{0.30}
      & {\footnotesize\textcolor{red}{(-0.09)}}\;\num{0.25}
      & {\footnotesize\textcolor{red}{(-0.06)}}\;\num{0.30}
      & {\footnotesize\textcolor{red}{(-0.06)}}\;\num{0.35}
      & {\footnotesize\textcolor{red}{(-0.09)}}\;\num{0.27}
      & {\footnotesize\textcolor{red}{(-0.06)}}\;\num{0.35} \\
   $\rightarrow$ Remove retriever
      & {\footnotesize\textcolor{red}{(-0.06)}}\;\num{0.32}
      & {\footnotesize\textcolor{red}{(-0.07)}}\;\num{0.27}
      & {\footnotesize\textcolor{red}{(-0.07)}}\;\num{0.30}
      & {\footnotesize\textcolor{red}{(-0.01)}}\;\num{0.39}
      & {\footnotesize\textcolor{red}{(-0.08)}}\;\num{0.29}
      & {\footnotesize\textcolor{ForestGreen}{(+0.01)}}\;\num{0.42} \\
   $\rightarrow$ Shuffle dataset
      & {\footnotesize\textcolor{red}{(-0.04)}}\;\num{0.34}
      & {\footnotesize\textcolor{red}{(-0.05)}}\;\num{0.29}
      & {\footnotesize\textcolor{red}{(-0.04)}}\;\num{0.32}
      & {\footnotesize\textcolor{ForestGreen}{(+0.00)}}\;\num{0.40}
      & {\footnotesize\textcolor{red}{(-0.05)}}\;\num{0.32}
      & {\footnotesize\textcolor{red}{(-0.00)}}\;\num{0.41} \\
   \bottomrule
\end{tabular}
  }
    \caption{\textbf{Both reasoning and retrieval are critical to \model{}'s performance.} We pick a random subset of users to perform ablations on. Removing the reasoning component leads to the largest drop in average performance ($-0.07$ absolute), while removing the retriever reduces performance by $-0.06$ on average. Shuffling the dataset also degrades results ($-0.04$), suggesting that preserving temporal structure is important. $u_{\mu}$ denotes mean performance across the evaluated users, and $u_i$ denotes performance on user $i$. We perform ablations on a random subset of 5 users.}
  \label{tab:ablations}
\end{table}

\begin{figure}[t]
    \centering
    \includegraphics[width=\linewidth]{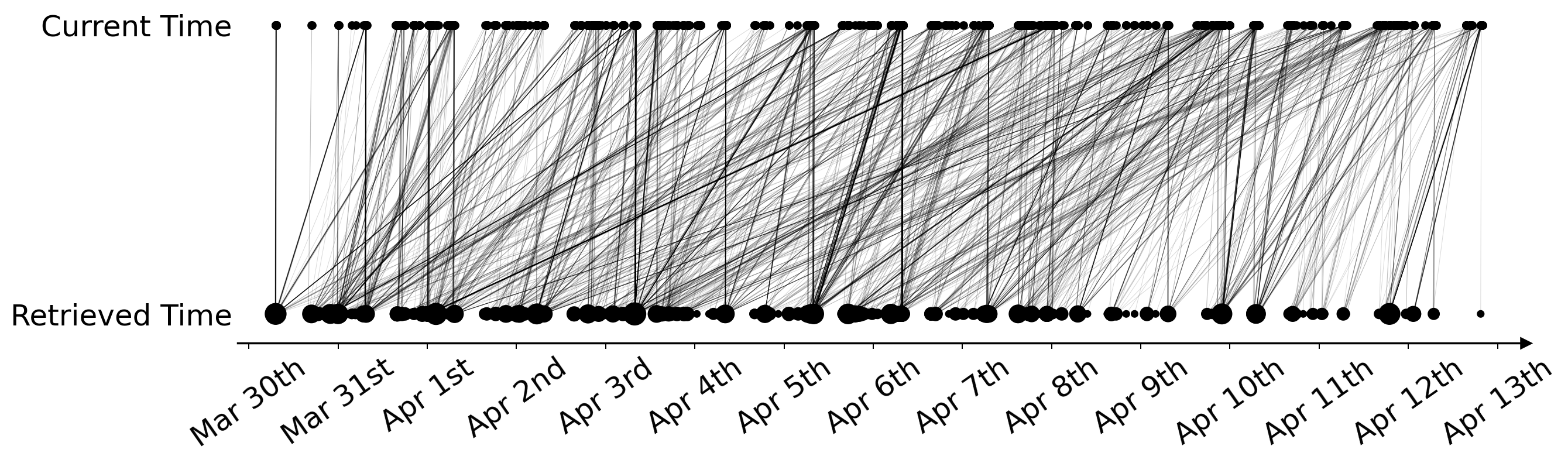}
    \caption{\textbf{At a given time (top), \model{} is likely to retrieve over a substantial part of its past context (bottom) to predict what a user will do next.} The visualization above (for a random user) shows what context from the past is retrieved for a query at the current time.}
    \label{fig:retrieved_analysis}
\end{figure}

In this section, we analyze the impact of various decisions in designing \model{} (\autoref{sec:ablations}) and analyze how reasoning traces evolve over the course of training and across users (\autoref{sec:reasoning_interp}).

\subsection{Algorithm Ablations} 
\label{sec:ablations}
We apply a handful of targeted ablations to \model{}, surfacing the impact of various components. First, we ablate \textbf{reasoning}: we optimize \model{} without generating reasoning traces for retrieval and prediction. Without reasoning, we retrieve only past observations, directly using the current observation as a query. In a separate ablation, we remove the \textbf{retriever} entirely, skipping the reasoning-to-retrieve step. Finally, we analyze the impact of our training order. We suspect chronological training over user traces helps model performance---reasoning traces accumulate and evolve in the order of observed interaction---so we \textbf{shuffle} our train data. To evaluate these ablations, we select a random subset of 5 users: the same subset of test users from our across-user generalization experiments (see \autoref{tab:many_user_results} and \autoref{sec:splits_exps}). 

All ablations reduce performance to varying degrees (main results in \autoref{tab:ablations}). First, removing reasoning degrades performance substantially (by 19.2\%, from 0.38 $\rightarrow$ 0.30). The same applies to retrieval: removing the ability to observe past reasoning and observations also degrades performance by 15.2\% (0.38 $\rightarrow$ 0.32). To illustrate the impact of the retriever, we additionally visualize how context is retrieved \emph{over} the course of training compared to the non-retriever ablation (\autoref{fig:retrieved_analysis}). \model{} learns to retrieve context from across its full interaction history, drawing on observations spread days apart rather than relying only on recent activity. Finally, shuffling the training data also has an impact (albeit smaller) on final performance: we observe a 9.3\% relative reduction in performance compared to \model{} (0.38 $\rightarrow$ 0.34).

\begin{figure}[t]
    \centering
    \includegraphics[width=\linewidth]{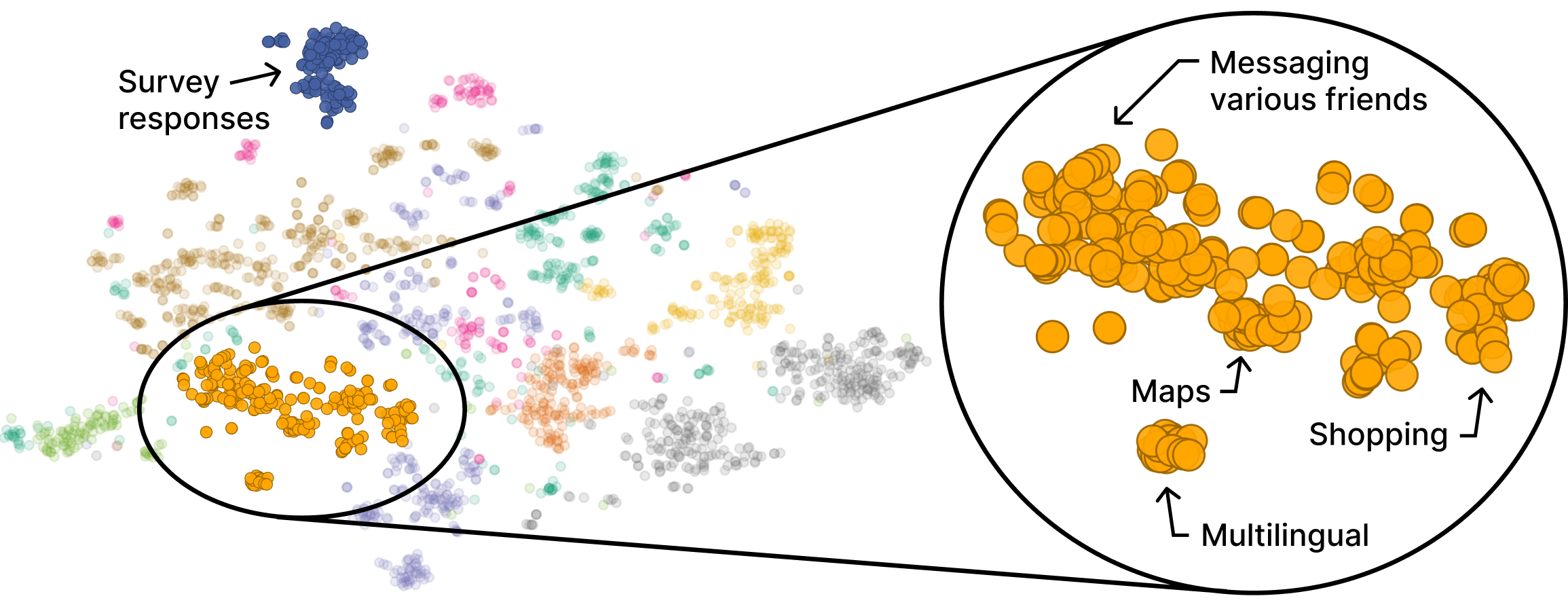}
    \caption{\textbf{\model{} learns diverse reasoning strategies when trained on different users.} Different colors correspond to different users. We embed (using the all-MiniLM-L6-v2 model; \citet{wang2020minilm,reimers2019sentence}) and visualize traces from the \emph{reasoning to predict} phase across a subset of users with tSNE. For some users (e.g. $u_7$, blue, almost exclusively takes online surveys), reasoning strategies are homogenous. For others, \model{} generates a library of reasoning patterns for specific contexts ($u_8$; in orange).}
    \label{fig:reasoning_viz}
\end{figure}

\subsection{Analyzing Reasoning Traces} 
\label{sec:reasoning_interp}
In our ablations, we find that allowing the model to reason plays a critical role in \model{}'s performance. In some cases, these traces may serve as explanations for a particular user's decisions~\citep{zhu2025using}. Here, we study how reasoning traces evolve during the course of training, across both reasoning for retrieval and for prediction. 

First, we analyze reasoning lengths from users for every epoch of training, for 10 full epochs. Across training, we find that reasoning traces generally get shorter, both for retrieval and prediction phases (\autoref{fig:cot_length}). Generally, traces for retrieval are shorter than for prediction (avg. 10.11 retrieval tokens v.s. 85.34 prediction). Thinking to retrieve traces become query-like (e.g. \trace{message Michael Diyi reminder} or \trace{ice cream salted caramel youtube}), likely optimized for the underlying BM25 retriever. 

Analyzing the content of the prediction traces themselves, we find substantial variance across---and often even within---users. Qualitatively, many of these traces describe a user's habits and preferences. To get a sense for this, we embed reasoning traces with sentence-transformers (using the all-MiniLM-L6-v2 model; \citet{wang2020minilm,reimers2019sentence}) and then visualize the embeddings (in \autoref{fig:reasoning_viz}). 

Some users have limited spread in reasoning traces. As a measure of spread, we compute the mean distance of all the user's embedded traces to the user centroid  ($r_\mathrm{avg}$). Consider the following trace from $u_7$, a user who uses their phone almost exclusively to complete surveys online ($r_\mathrm{avg}$ = 4.94): 
\begin{quote}
    \trace{The user's actions suggest a habitual and possibly patterned behavior of completing surveys and managing their Yahoo Mail inbox. This could involve clicking "Next” on a survey to "Continue,” possibly indicating the completion of a set of survey questions...}
\end{quote}
Many of this user's traces are similar in content and style. In contrast, other users have models that produce more diverse traces, capturing variance in their behavior. $u_8$ ($r_\mathrm{avg}$ = 16.26) regularly goes house-hunting during the day:
\begin{quote}
    \trace{The user systematically goes through an interactive map, view detailed photos of houses on the listing details page, and then pan again through the [ANONYMIZED] area properties.}
\end{quote}
and switches between various social media apps at night:
\begin{quote}
    \trace{Since communication on personal social media platforms is a common activity, following the successful completion of Instagram Stories, the user may turn to other notifications to check for any incoming emails, messages from their list of contacts, or potential bottom notifications from social media apps.}
\end{quote}
While reasoning improves our models' predictive power, exploring if the reasoning itself is indeed faithful or accurate is an avenue for future work.
\section{Beyond Next Action Prediction}
We outline two applications enabled by our work; namely the ability to learn entirely online from user interactions, and the ability to generalize as helpful assistants. 

\paragraph{Learning online} The distribution of a person's behavior is always shifting. As the person changes over time, so does the work they do and the patterns they exhibit. LongNAP should be able to adapt to this drift as we observe additional interaction data. 

In our work, we present data collection and training as two processes that occur \emph{synchronously}, but this need not be the case! Instead of storing data and training offline with multiple epochs, we can convert the entire pipeline to run online, where training proceeds continuously in the background; or overnight \citep{lin2025sleep}. We call this version powerNAP. In powerNAP, NAPsack and LongNAP operate asynchronously: NAPsack continuously tracks and labels user actions, enqueueing them for training, while LongNAP consumes labeled actions from the queue and trains on them in a single pass, discarding data after use. Crucially, memory is never reset and reasoning traces accumulate, allowing the model to continually build a better representation of the user over time. We release powerNAP as a demo\footnote{Available at \url{https://github.com/GeneralUserModels/powernap}} for users to try on their own data.

\paragraph{Assistants} Once a user has a good \model{}, we can anticipate what they want and intend to do. This should enable an assistant that finishes predictable tasks for users by acting on predictions about what the user would do next. While \model{} is not a computer use agent, we can easily use it to pilot one. We release a simple version of such an assistant, called SleepWalk, which relies on an off-the-shelf computer use agent~\citep{anthropic2024computeruse} to execute on actions predicted by LongNAP.

\section{Related Work}

\paragraph{World, Human, and User Models} The ability to predict the dynamics of complex physical and social behaviors are longstanding goals in both human-computer interaction and artificial intelligence. Progress on large, multimodal models has revitalized this vision. Trained on enough video data, large-scale generative video models show promise in \emph{predicting} world dynamics, for example through next frame prediction~\citep{hafner2019dream, bruce2024genie, yang2023learning}. These predictive world models have opened a range of research directions in robotics, enabling data efficient robotic learning~\citep{sharma2023towards, quevedo2025worldgym, du2023learning}. Similarly, LLMs have been used to simulate general human behavior~\citep{park2023generative, wu2026humanlmsimulatingusersstate} for social science research~\citep{argyle2023out, hewitt2024predicting, park2024generative} or to build proactive, question-asking assistants~\citep{sun2025training, wu2025collabllm}. Both approaches rely on a similar assumption: that the internet contains substantial amounts of realistic behavioral data to bootstrap simulation. While these methods are sometimes effective, they tend to suffer from a \emph{sim-to-real} gap~\citep{zhao2020sim}. In other words, we apply these simulators in very specific situations (e.g. specific individuals, robots, etc.) that are out of distribution. And continually collecting new training data to mend this distribution shift is technically challenging~\citep{mirchandani2024so} and/or prohibitively expensive~\citep{halevy2009unreasonable}. 

In this paper, we train models directly at the level of an individual user---a user model. The user models cannot afford to suffer from the sim-to-real gap: they are immediately deployed to specific people. So instead of relying on datasets that serve as proxies, we directly collect data from individual interaction traces. We rely on work showing that VLMs can effectively describe individual behavior through observation~\citep{shaikh2025creating, wang2025ai}. We then validate and build infrastructure (\autoref{sec:data_collection}) for continually labeling low-level trajectories of individual behavior at scale. 

\paragraph{Personal Reasoning} Training models that understand what people want requires \emph{personal} reasoning~\citep{li2025personalized}: the ability to flexibly reason over our opaque personal preferences and beliefs. In contrast, the predominant RL setting relies on tasks where the outcome is easily verifiable, like math or symbolic reasoning~\citep{silver2016mastering, silver2017mastering, trinh2024solving}. Most LLMs are similarly trained to reason on easily verifiable tasks~\citep{zelikman2022star, cobbe2021training}. Because of this reliance on verifiability, LLM reasoning often fails to generalize beyond these domains~\citep{shaikh2023second, sprague2024cot}. Training a model that can instead reason effectively over everyday interaction enables a range of human-centered applications: from proactive assistants that know your context well enough to autonomously do ``the right thing at the right time,''~\citep{weiser1991computer} to AI models that know when and how to defer effectively to users~\citep{horvitz1999principles}.

In our work, we train such a model (\model{}) over everything a person sees and does on their computer. Most related to our work is \citet{gandhi2026learning}, where an LLM is trained to reason for dialogue simulation. We instead learn to predict a user's general actions over the entirety of their digital context. In addition, \model{} takes inspiration from work on metacognitive reuse~\citep{suzgun2025dynamic, didolkar2025metacognitive, sarukkai2025self} and reasoning abstractions~\citep{qu2025rlad}, where reasoning traces are re-used over the course of learning. Likewise, \model{} learns to both generate, retrieve, and re-use reasoning traces at the \emph{individual} user level.

\paragraph{Memory and Retrieval} 

To predict a user's next action, we must be able to use the entirety of their digital context, which can span months, years, or more. However, LLMs have practical context limitations. The context window is constrained to a finite number of tokens; and putting everything in-context can degrade model performance~\citep{liu2024lost}. One solution involves retrieval-augmented generation, where LLMs retrieve from a larger, external database~\citep{chen2017reading, lewis2020rag}. Instead of jointly optimizing a dense retriever end-to-end, LLMs can also be trained to use retrievers by generating queries to retriever \emph{tools}~\citep{schick2023toolformer, hsu2024grounding, jin2025search}. In addition, LLMs can judge the relevance of retrieved context for question-answering tasks~\citep{asai2024self}. 

Similarly, we introduce models that can effectively retrieve context for next action prediction, querying a lexical retriever. Our setting is unique in a few ways. First, we are retrieving relevant context not from an external search index or a database, but from the user's own interaction history. Second, we \emph{reason to retrieve} what context is relevant specifically for predicting a user's next action. This process is end-to-end optimized via policy gradient methods; and significantly improves \model{} performance. 
\section{Discussion}
\label{sec:disc_limit}

In our evaluations, we find that \model{}s show promise in predicting what users will do next across their digital contexts. We discuss implications of deploying models like \model{} from privacy and alignment perspectives, and outline avenues for future work.

\paragraph{Privacy} Models like \model{} operate over large swaths of our context. Inevitably, they will contain private and sensitive data about users. Our architecture does limit some of this exposure, since learning to retrieve keeps the traces local; at some performance cost, one could either build the entire model locally or ensure that the learning to reason stage is not finetuned on any specific user. In addition, in our work, we rely on an approved infrastructure for processing personally identifiable information (PII) and personal health data (PHI). At our institution, only Google Cloud services are approved for processing PHI; so we rely on vetted, private pipelines to access Google's Gemini models for annotation. All model training occurs on open models (Qwen-2.5-VL-7B), where compute instances are managed by the research team.

We recognize that these precautions are very challenging to take for the individual user. The \emph{privacy paradox}~\citep{norberg2007privacy} makes deploying models like \model{} in a centralized fashion difficult. In other words, users are likely to disclose more to \model{}, especially given (1) the ease of collecting data and (2) the benefits that come with a proactive AI system.  

There are several promising approaches that can mitigate these privacy concerns. The first is decentralization. We suspect that models will continue to get cheaper and faster, enabling on-device inference and training. Methods like FlashAttention~\citep{dao2022flashattention}, effective quantization~\citep{dettmers2023qlora}, or specialization via synthetic data~\citep{shen2026sera} already save substantially on compute or memory. If local models remain difficult, we can still redact private information with a smaller local model~\citep{li2025papillon}, only share private data to a larger model~\citep{nissenbaum2004privacy, mireshghallah2023can,  shao2024privacylens} based on a user's personal context~\citep{shaikh2025creating}, or decouple model requests from eachother through a VPN-like system~\citep{liu2026unlinkable}.

\paragraph{Aligning \model{}s} In our current instantiation of \model{}, we train models to do what a user might do next. There are many instances where this may not be helpful. For example, a user who habitually procrastinates may not want to use a model that helps them procrastinate. This is a challenging alignment problem with parallels to both filter bubbles on social media~\citep{pariser2011filter, munson2010presenting, bakshy2015exposure} and sycophancy in chat-based LLMs~\citep{cotra2021ai, perez2023discovering, cheng2025social}. We want learned \model{}s to complement users in ways that help. A promising avenue for future work involves applying methods for eliciting values to steer social media algorithms~\citep{popowski2026socialmediafeedelicitation}. Similar methods could be applied to \model{}s.

\paragraph{Limitations and Future Work} There are fundamental limitations to learning from just observation. In our setting, models will only be able to make inferences from what happens on a user's screen, which is still a narrow proxy for a user's general context~\citep{dourish2004we}. We are still far from models that can draw from everyday action beyond our devices, but we suspect our approach can be generalized to interaction beyond screenshots. 

Both our labeling and training processes rely on large pretrained models. First, our training data itself is generated by a VLM captioning user activity. We find that VLMs are fairly performant at this task, and we expect performance to improve as VLMs improve. Still, they are not perfect---errors in captioning will cascade down to training and prediction. Like prior work, we also rely on LLM-as-a-judge for both reward and optimization~\citep{bai2022constitutional, dubois2023alpacafarm}. We find that this metric continues to correlate with human judgement---humans pick samples from \model{} over all other training approaches (\S\ref{sec:overall_results}). For longer runs, however, the judge alone may be prone to reward hacking \citep{wang2025thinkingcheatingdetectingimplicit, gandhi2026learning}. We leave experimenting with other rewards for future work.

We also presented a basic scaffold for training \model{}s. This scaffold could be made more expressive by allowing LongNAP to interleave retrieval and reasoning within a single generation pass, and by equipping it with additional tools such as web search. While we validated our method on a small sample of users and showed that LongNAP generalizes both over time and across users, it would be valuable to study how performance scales over longer time horizons and with many more users. Training separate weights for every user also presents practical challenges. Future work could explore how to efficiently train and serve per user LoRAs at scale \citep{sheng2023s,chen2024punica}. Further, while current mid-training and pretraining data \citep{olmo2025olmo,havrilla2024surveying,penedo2024fineweb} for LLMs are optimized to improve performance on science, code and math, one can imagine that other types of data and reasoning strategies \citep{gandhi2025cognitive} could be better for NAP. 

Finally, a few training limitations. First, we experiment only with GRPO~\citep{shao2024deepseekmath} as our policy gradient objective, due to the added memory constraints of learning an entire value network for methods like PPO~\citep{schulman2017proximal}. In addition, we were unable to train all models to convergence because of budget constraints (our validation scores from Fig.~\ref{fig:val_scores_plot} continue to increase, for example). We suspect that performance estimates in this paper may be a lower bound, and leave continued training experiments to future work.

\section{Conclusion}
We introduced LongNAP, a long-context next action predictor that learns to anticipate what users will do next by reasoning over their full multimodal interaction history. To collect training data, we introduced NAPsack, a passive pipeline that annotates naturalistic behavior traces at scale using vision-language models---demonstrating that rich, labeled interaction data can be obtained without any active user effort. In evaluations across 20 users and 1,800 hours of screen time, LongNAP significantly outperforms supervised finetuning and prompted baselines when trained on individual users. We also observe modest generalization when training on many users and generalizing to new ones. Altogether, we argue that learning from the full context of user behavior to anticipate user needs is now a tractable direction.

\section*{Contribution Statement} OS conceived the initial idea, and planned/evaluated all experiments. VT and KG helped develop data labeling code, proposed and tested critical modeling ideas, and helped with framing the paper. KG also built an online implementation (powerNAP) of the paper. YC helped with Screenomics infrastructure, labeling, and data collection. DY, MB, and SY were primary co-supervisors for this project. All authors discussed results and contributed in writing the final paper.

\section*{Acknowledgements}

We thank Dora Zhao, Michael Li, Vindula Jayawardana, Shardul Sapkota, Matthew Jörke, Helena Vasconcelos, Gerard de Melo, Michelle Lam, Shan Rizvi, Chris Rytting, and Vishnu Sarukkai for helpful discussions and feedback. Omar Shaikh is supported by the HAI-HPI program. The Screenomics components of this study were supported in part by a grant from the National Heart, Lung, and Blood Institute of the National Institutes of Health, under Award number R01HL16901. The content is solely the responsibility of the authors and does not necessarily represent the official views of the National Institutes of Health or other funders. Finally, we appreciate the support from Sloan Foundation, Laude Institue, Thinking Machines (for Tinker credit), and Stanford Institute for Human-Centered Artificial Intelligence, as well as ONR grant N00014-24-1-2532.

\bibliography{ref}
\bibliographystyle{colm2026_conference}
\clearpage
\newpage

%%%%%%%%%%%%%%%%%%%%%%%%%%%%%%%%%%%%%%%%%%%%%%%%%%%%%%%%%%%%

\appendix

\section{NAPsack}
\label{appendix:napsack}

To record a session with NAPsack and compare it to a baseline without event-driven compression, we implement current active screen capturing using ffmpeg and apply the same recording hyperparameters for both methods.

\subsection{Hyperparameters}
\label{appendix:pack_hyperparams}

NAPsack uses thresholds to group input events and decide when screenshots should be persisted. All recordings are performed at 30 FPS and a resolution of 1920$\times$1080. To ensure that interface states immediately before and after interactions are preserved, NAPsack stores screenshots 75ms before the first event of a burst and 75ms after its last event.

\begin{table}[h]
\centering
\begin{tabular}{lcc}
\toprule
Event type & GAP threshold (s) & MAX duration (s) \\
\midrule
Click  & 0.2 & 0.3 \\
Move   & 0.5 & 4.0 \\
Scroll & 0.5 & 3.0 \\
Key    & 0.5 & 6.0 \\
\bottomrule
\end{tabular}
\caption{Event burst thresholds used by NAPsack.}
\label{appendix:napsack_thresholds}
\end{table}

\subsection{Grouping Nearby Events into Bursts}
\label{appendix:napsack_bursts}

NAPsack groups temporally adjacent input events of the same type into \emph{event bursts}. An event is assigned to the current burst if the time since the preceding event of that type does not exceed the corresponding \textbf{gap} threshold and the elapsed time since the burst start remains within the \textbf{max} duration (see table~\ref{appendix:napsack_thresholds}). If the \textbf{gap} threshold is exceeded, a new burst is started. If the \textbf{max} duration is exceeded, the first half of the current burst is finalized and saved, while the second half becomes the active burst. A burst is force-restarted when the active monitor changes. All thresholds were determined qualitatively; and should be re-tuned for new interfaces. 

\subsection{Label Prompts}
\label{appendix:pack_prompts}

We include all prompts for NAPsack in the following repo: \url{https://github.com/GeneralUserModels/napsack/tree/main/src/label/prompts}

\begin{figure}
    \centering
    \includegraphics[width=\linewidth]{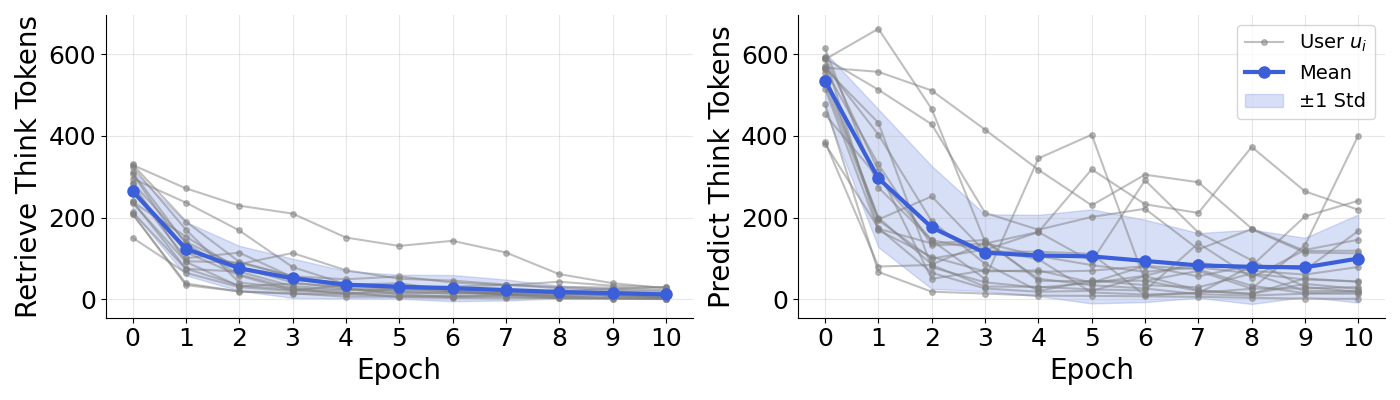}
    \caption{\textbf{Reasoning traces grow shorter across model training.} Traces for the retrieve phase grow far shorter than queries for the predict phase (avg. 10.11 retrieval tokens v.s. 85.34 prediction). Qualitatively, we find that reasoning for the retrieve phase resembles queries to a retriever; while reasoning for the prediction phase resembles higher order descriptions of a user's behavior.}
    \label{fig:cot_length}
\end{figure}

\subsection{Annotating Ground Truth Labels for NAPsack}
\label{appendix:pack_gt}

Both authors verified ground truth labels recorded from personal screen recordings. To construct ground truth labels, the authors selected the best outputs generated from each NAPsack condition, and manually corrected the trajectories to match ground truth. Both authors reviewed the ground truth trajectories and preference annotations for errors over discussion. The author from whom the recordings were sourced resolved mistakes in ground truth labels during annotation.

\subsection{Annotating Screenomics}
\label{appendix:screenomics_users}

\paragraph{Demographics} The Screenomics dataset we use comes from \cite{reeves2021screenomics}. We subsampled 20 participants, of which one had no demographic data. The remaining 19 participants (14 female, 5 male) were located across the United States and ranged in age from 22 to 70, with a mean of 44 and a median of 39. The majority identified as White (16/19), with 2 identifying as Asian and 1 as Black; 6 identified as Hispanic. Education levels varied across participants. 2 held a high school diploma, 7 had some college, 1 an associate's degree, 5 a bachelor's, and 4 a graduate degree.

\paragraph{Deduplicating Images} Many images in the Screenomics dataset are duplicates. To identify and remove screenshots where screen content is unchanged, we compute a perceptual difference hash for each image~\citep{krawetz2011dhash}. This works by resizing the image to a small fixed size, comparing adjacent pixel intensities to produce a fingerprint, and then measuring Hamming distance between fingerprints of consecutive screenshots. Qualitatively, pairs with a distance at or below a threshold of 5 (out of a 16x16 = 256-bit hash) are near-duplicates; so we filter these images. 

\subsection{Judge Prompt}
\label{appendix:judge_examples}

We include our judge prompt for training here: \url{https://github.com/GeneralUserModels/powernap/blob/main/src/powernap/longnap/verifiers/accuracy.txt}
\section{\model{} details}

\begin{table}
\centering
\begin{tabular}{l|rr}
\toprule 

 Hyperparameter & \model{} & SFT \\
 \midrule
 Effective Batch Size & 16 & 16 \\
 Learning Rate &  3e-5 & 1e-4 \\
 Group Size & 4 & - \\
 Epochs & 10 & 10 \\
 Actions in history & 16 & 16 \\
 Actions to predict & 8 & 8 \\
 Images in history & 2 & 2 \\
 LoRA rank & 8 & 8 \\
 \phantom{LoRA }alpha & 32 & 32\\
 \phantom{LoRA }dropout & 0.05 & 0.05\\
 \phantom{LoRA }modules & MLP only & MLP only\\
\bottomrule
\end{tabular}
\caption{\textbf{Hyperparameter settings.} We the best checkpoint based on our validation performance, early stopping if appropriate. We only sweep learning rate on baselines (with SFT) due to budget constraints. We train on 8 B200 GPUs.}
\label{tab:hparams_models}
\end{table}

\begin{figure}[t]
    \centering
    \includegraphics[width=0.6\linewidth]{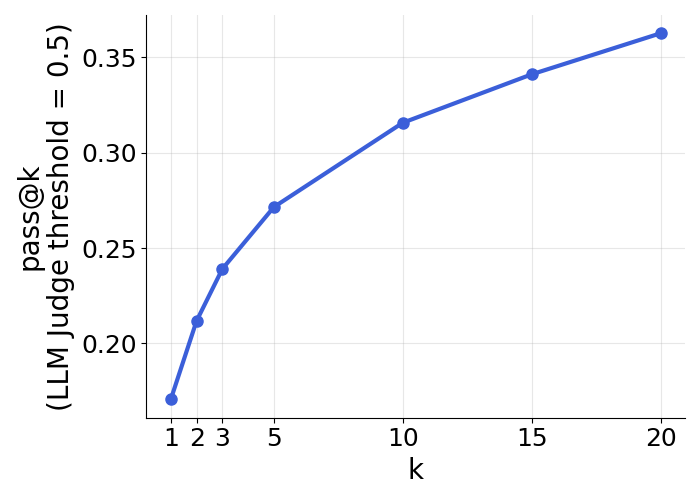}
    \caption{\textbf{Pass@k scores for \model{}.} To count as a ``pass,'' we selected an LLM-judge threshold of 0.5; trajectories that get this score are often well aligned with the actual intent of the user, but miss minor details or skip a few actions (see \autoref{sec:metrics} for an example). At this threshold, LongNAP achieves \textbf{17.1\% at pass@1} across users, rising to \textbf{36.3\% at pass@20}.}
    \label{fig:pass_at_k}
\end{figure}

\begin{figure}[t]
    \centering
    \includegraphics[width=1\linewidth]{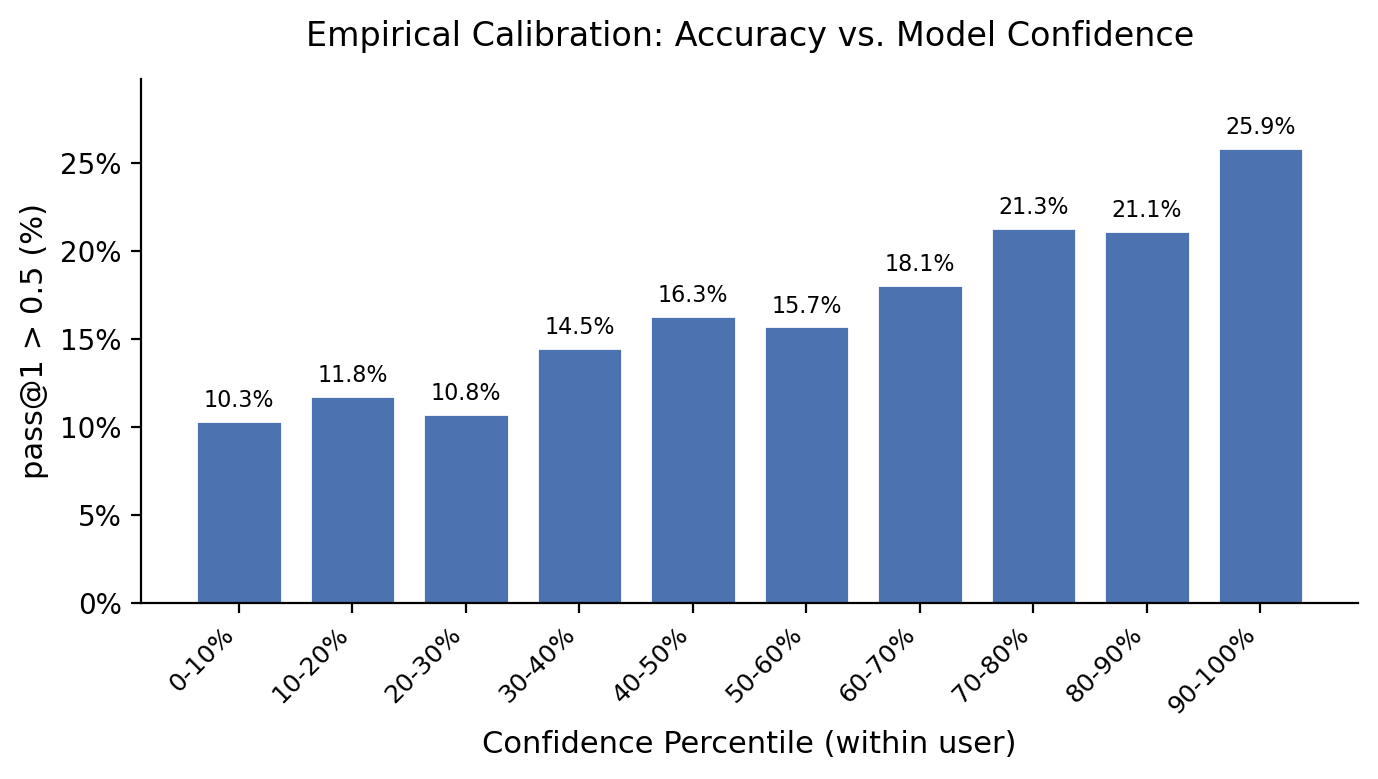}
    \caption{\textbf{Empirical calibration of LongNAP.} Confidence percentiles are computed per-user from intra-cluster variance of 20 sampled trajectories. Higher confidence prompts yield substantially higher pass@1 accuracy (25.9\% vs. 10.3\%).}
    \label{fig:empiric_calib}
\end{figure}

\subsection{Training and Retriever Hyperparamters}
\label{appendix:longnap_hyperparameters}
\paragraph{Training} We include hyperparameter details for \model{} and our SFT baseline in \autoref{tab:hparams_models}. We detail how many actions we place in context; how many we predict; and the number of images placed in context. Even with LoRA (hyperparameters also in \autoref{tab:hparams_models}), we find that images significantly increase contexts lengths. Because of memory limitations, we only keep images associated with the last 2 events in context. Across all models, we sample with temperature 1.0.

% \paragraph{Retriever}
\paragraph{Retriever} \model{} uses an in-memory BM25 retriever to retrieve old reasoning traces and observations. To start, we use standard BM25 parameters $k_1 = 1.5$ and $b = 0.75$. In addition, we apply a temporal decay. At query time, each candidate's BM25 score is multiplied by $\exp(-\lambda \cdot \text{age})$, where, $\text{age}$ is the time difference between the query and the document's event measured in days. We set $\lambda = 0.5$, which strongly favors recent context over older history. We also apply a constraint to make sure that retrieved items are diverse. We first retrieve the top-$k = 10$ candidates from BM25, then apply Maximal Marginal Relevance (MMR) reranking to select $k_{\text{mmr}} = 5$ results with diversity parameter $\alpha = 0.5$, balancing relevance and diversity equally~\citep{carbonell1998use}. Finally, to avoid redundancy in the retriever's memory, we deduplicate when we insert. We use trigram Jaccard similarity (threshold of $0.8$), replacing older near-duplicate entries with newer ones. 

\subsection{Prompting Baselines}
\label{appendix:prompt_baselines}

For prompting baselines, we try zero-shot and few-shot prompting. Few-shot prompts use our retriever (see details in ~\autoref{appendix:longnap_hyperparameters} to retrieve and place relevant few-shot examples in-context. Below, we outline our prompt (few-shot additions are {\color{red}highlighted in red}), used for both closed and open source models:

\begin{tcolorbox}[title=Prompt,
colback=gray!5!white,
colframe=black,
fonttitle=\bfseries,
coltitle=white,
breakable,
enhanced jigsaw]
You are an expert at analyzing user behavior patterns and predicting future actions based on historical context.

\vspace{1em}

\#\# Task

Analyze the following sequence of user actions and predict what they will do next. Pay attention to:

- The temporal patterns and timing of actions

- The logical flow and context of the user's workflow

- Common behavioral patterns that might indicate intent

\vspace{1em}

\#\# Previous Actions

\texttt{\textless actions\textgreater}

\texttt{\{past\_16\_actions\}}

\texttt{\textless /actions\textgreater}
\vspace{1em}

{\color{red}
\#\# Relevant Retrieved Actions

\texttt{\textless actions\textgreater}

\texttt{\{retrieved\_actions\}}

\texttt{\textless /actions\textgreater}
}
\vspace{1em}

\#\# Instructions

Based on the actions above, predict the most likely next **8** actions the user will take. Output your predictions in the following format:

\vspace{1em}

\texttt{\textless actions\textgreater}

\texttt{\{example\_actions\}}

\texttt{\textless /actions\textgreater}

\vspace{1em}

Be specific and realistic in your predictions. Consider the user's apparent goals and typical workflows. You MUST output exactly 8 actions.
\end{tcolorbox}

 \begin{table*}[t]
  \centering
  \small
  \caption{Qualitative examples of predicted vs.\ ground-truth actions at different accuracy scores assigned by the Gemini judge (Part 1: low scores). The first five rows are from model predictions.}
  \label{tab:judge-examples-1}
  \begin{tabularx}{\textwidth}{c p{0.38\textwidth} p{0.38\textwidth}}
  \toprule
  \textbf{Score} & \textbf{Ground Truth} & \textbf{Predicted Actions} \\
  \midrule
  % --- 0.05: completely wrong app and intent ---
  0.05 &
  \begin{minipage}[t]{\linewidth}\raggedright\scriptsize
  1. Typed ``icc world cup'' into search bar \\
  2. Clicked ``STANDINGS'' tab in ICC T20 World Cup widget \\
  3. Scrolled through standings table \\
  4. Clicked ``SL'' (Sri Lanka) in standings \\
  5. Switched to W\&B dashboard tab \\
  6. Scrolled through W\&B charts \\
  7. Clicked ``train/entropy'' chart in W\&B \\
  8. Clicked chat with ``Omar Shaikh'' in Slack
  \end{minipage}
  &
  \begin{minipage}[t]{\linewidth}\raggedright\scriptsize
  1. Typed ``x'' into address bar, pressed Enter \\
  2. Clicked Twitter logo in top left \\
  3. Typed ``arch'' into Twitter search bar \\
  4. Clicked search result for ``Arch Labor'' \\
  5. Clicked ``Home'' button in Twitter sidebar \\
  6. Scrolled through Twitter feed \\
  7. Clicked search button on Twitter \\
  8. Typed ``forms'' into Twitter search bar
  \end{minipage} \\
  \midrule
  % --- 0.12: right topic, wrong specific actions ---
  0.12 &
  \begin{minipage}[t]{\linewidth}\raggedright\scriptsize
  1. Clicked ``rouge-score'' on PyPI, copied install command \\
  2. Ran \texttt{uv pip install rouge-score} in terminal \\
  3. Scrolled through ``rouge-score'' PyPI page \\
  4. Ran \texttt{accelerate launch} training command \\
  5. Scrolled through ``rouge-score'' PyPI page \\
  6. Switched between W\&B, Google Doc, and terminal \\
  7. Typed training command, cancelled with Ctrl+C \\
  8. Ran \texttt{cd sim-v2} and \texttt{git pull}
  \end{minipage}
  &
  \begin{minipage}[t]{\linewidth}\raggedright\scriptsize
  1. Scrolled through search results for ``install rouge score'' \\
  2. Clicked on a search result for ``install rouge score'' \\
  3. Scrolled down ``crag'' repository on GitHub \\
  4. Clicked ``hub/parouza'' link in GitHub repo \\
  5. Clicked ``eval'' directory in file explorer \\
  6. Pressed Enter to run \texttt{evaluate\_qm9.py} \\
  7. Switched to Google Chrome window \\
  8. Copied \texttt{pip install} command from GitHub
  \end{minipage} \\
  \midrule
  % --- 0.20: right app, wrong interactions ---
  0.20 &
  \begin{minipage}[t]{\linewidth}\raggedright\scriptsize
  1. Clicked on a purple property pin on the map \\
  2. Clicked on the ``\$3.5k+'' property pin \\
  3. Clicked on the ``\$2.8k'' property pin \\
  4. Clicked on the ``\$3.5k+'' property pin \\
  5. Clicked on the ``\$3.3k'' property pin \\
  6. Clicked on the ``\$3.9k'' property pin \\
  7. Clicked on the ``\$5.3k'' property pin \\
  8. Clicked on the ``\$2.8k'' property pin
  \end{minipage}
  &
  \begin{minipage}[t]{\linewidth}\raggedright\scriptsize
  1. Scrolled through search results and map on Zillow \\
  2. Clicked on a property marker on the Zillow map \\
  3. Clicked ``Down'' arrow to expand price filter \\
  4. Clicked ``OK'' to apply price filter settings \\
  5. Scrolled through filtered search results \\
  6. Selected a property listing card from results \\
  7. Clicked ``Show on map'' on the listing page \\
  8. Scrolled through rental listings on Zillow map
  \end{minipage} \\
  \midrule
  % --- 0.35: right apps, wrong sequence ---
  0.35 &
  \begin{minipage}[t]{\linewidth}\raggedright\scriptsize
  1. Scrolled through W\&B dashboard \\
  2. Clicked ``kanishkgandhi'' project link in W\&B \\
  3. Clicked on a run in W\&B sidebar \\
  4. Clicked ``kanishkgandhi's workspace'' tab \\
  5. Scrolled through Claude conversation \\
  6. Clicked ``New Tab'' button in Chrome \\
  7. Switched to W\&B tab \\
  8. Scrolled through W\&B for ``context-distillation-goyal''
  \end{minipage}
  &
  \begin{minipage}[t]{\linewidth}\raggedright\scriptsize
  1. Scrolled through W\&B charts on ``NAP Time'' dashboard \\
  2. Closed W\&B dashboard tab in Chrome \\
  3. Switched to Claude.ai tab \\
  4. Closed current tab in Chrome \\
  5. Switched to Claude tab \\
  6. Inspected text in Claude chat \\
  7. Re-fetched view selection content from Claude \\
  8. Re-fetched view selection content from Claude
  \end{minipage} \\
  \midrule
  % --- 0.42: right content, partially matching ---
  0.42 &
  \begin{minipage}[t]{\linewidth}\raggedright\scriptsize
  1. Switched to ``Australia vs Zimbabwe'' tab, entered fullscreen \\
  2. Clicked thumbnail in ``Key Moments'' section \\
  3. Pressed ``esc'' to exit fullscreen \\
  4. Switched to W\&B workspace, clicked ``Runs'' icon \\
  5. Switched to ``Australia vs Zimbabwe,'' entered fullscreen \\
  6. Pressed ``esc'' to exit fullscreen \\
  7. Switched to W\&B tab, scrolled through charts \\
  8. Scrolled through W\&B dashboard
  \end{minipage}
  &
  \begin{minipage}[t]{\linewidth}\raggedright\scriptsize
  1. Switched to ``Australia vs Zimbabwe'' tab \\
  2. Clicked full screen button on cricket video \\
  3. Clicked History channel logo at bottom of player bar \\
  4. Pressed ``escape'' in Chrome \\
  5. Clicked on the video player \\
  6. Clicked History channel logo at bottom of player bar \\
  7. Clicked History channel logo, pressed ``space'' \\
  8. Pressed ``space'' in Chrome
  \end{minipage} \\
  \bottomrule
  \end{tabularx}
  \end{table*}

  \begin{table*}[t]
  \centering
  \small
  \caption{Qualitative examples continued (Part 2: high scores). These examples are manually constructed to illustrate what the upper score range represents.}
  \label{tab:judge-examples-2}
  \begin{tabularx}{\textwidth}{c p{0.38\textwidth} p{0.38\textwidth}}
  \toprule
  \textbf{Score} & \textbf{Ground Truth} & \textbf{Predicted Actions} \\
  \midrule
  % --- 0.52: right pattern, wrong details in ~half ---
  0.52 &
  \begin{minipage}[t]{\linewidth}\raggedright\scriptsize
  1. Switched to ``Australia vs Zimbabwe'' tab, entered fullscreen \\
  2. Clicked thumbnail in ``Key Moments'' section \\
  3. Pressed ``esc'' to exit fullscreen \\
  4. Switched to W\&B workspace, clicked ``Runs'' icon \\
  5. Switched to ``Australia vs Zimbabwe,'' entered fullscreen \\
  6. Pressed ``esc'' to exit fullscreen \\
  7. Switched to W\&B tab, scrolled through charts \\
  8. Scrolled through W\&B dashboard
  \end{minipage}
  &
  \begin{minipage}[t]{\linewidth}\raggedright\scriptsize
  1. Switched to ``Australia vs Zimbabwe'' tab, clicked player \\
  2. Pressed ``f'' to enter fullscreen on cricket video \\
  3. Pressed ``esc'' to exit fullscreen \\
  4. Opened new tab, typed ``wandb.ai'' into address bar \\
  5. Scrolled through W\&B charts for training metrics \\
  6. Clicked ``Table'' view in W\&B dashboard \\
  7. Switched to YouTube, clicked recommended video \\
  8. Scrolled through comments section on YouTube
  \end{minipage} \\
  \midrule
  % --- 0.72: first half exact, second half diverges ---
  0.72 &
  \begin{minipage}[t]{\linewidth}\raggedright\scriptsize
  1. Switched to ``Australia vs Zimbabwe'' tab, entered fullscreen \\
  2. Clicked thumbnail in ``Key Moments'' section \\
  3. Pressed ``esc'' to exit fullscreen \\
  4. Switched to W\&B workspace, clicked ``Runs'' icon \\
  5. Switched to ``Australia vs Zimbabwe,'' entered fullscreen \\
  6. Pressed ``esc'' to exit fullscreen \\
  7. Switched to W\&B tab, scrolled through charts \\
  8. Scrolled through W\&B dashboard
  \end{minipage}
  &
  \begin{minipage}[t]{\linewidth}\raggedright\scriptsize
  1. Switched to ``Australia vs Zimbabwe'' tab, entered fullscreen \\
  2. Clicked on video player timeline to skip forward \\
  3. Pressed ``esc'' to exit fullscreen \\
  4. Switched to W\&B workspace, scrolled through dashboard \\
  5. Switched to ``Australia vs Zimbabwe'' tab, entered fullscreen \\
  6. Pressed ``esc'' to exit fullscreen \\
  7. Switched to W\&B tab, clicked on a chart \\
  8. Clicked ``Table'' view in W\&B workspace
  \end{minipage} \\
  \midrule
  % --- 0.78: 6/8 exact, 2 wrong prices ---
  0.78 &
  \begin{minipage}[t]{\linewidth}\raggedright\scriptsize
  1. Clicked on a purple property pin on the map \\
  2. Clicked on the ``\$3.5k+'' property pin \\
  3. Clicked on the ``\$2.8k'' property pin \\
  4. Clicked on the ``\$3.5k+'' property pin \\
  5. Clicked on the ``\$3.3k'' property pin \\
  6. Clicked on the ``\$3.9k'' property pin \\
  7. Clicked on the ``\$5.3k'' property pin \\
  8. Clicked on the ``\$2.8k'' property pin
  \end{minipage}
  &
  \begin{minipage}[t]{\linewidth}\raggedright\scriptsize
  1. Clicked on a purple property pin on the map \\
  2. Clicked on the ``\$3.5k+'' property pin \\
  3. Clicked on the ``\$2.8k'' property pin \\
  4. Clicked on the ``\$3.5k+'' property pin \\
  5. Clicked on the ``\$3.3k'' property pin \\
  6. Clicked on the ``\$3.9k'' property pin \\
  7. Clicked on the ``\$4.2k'' property pin \\
  8. Clicked on the ``\$3.1k'' property pin
  \end{minipage} \\
  \midrule
  % --- 0.82: 6/8 exact, 2 differ in named entities ---
  0.82 &
  \begin{minipage}[t]{\linewidth}\raggedright\scriptsize
  1. Typed ``icc world cup'' into search bar \\
  2. Clicked ``STANDINGS'' tab in ICC T20 World Cup widget \\
  3. Scrolled through standings table \\
  4. Clicked ``SL'' (Sri Lanka) in standings \\
  5. Switched to W\&B dashboard tab \\
  6. Scrolled through W\&B charts \\
  7. Clicked ``train/entropy'' chart in W\&B \\
  8. Clicked chat with ``Omar Shaikh'' in Slack
  \end{minipage}
  &
  \begin{minipage}[t]{\linewidth}\raggedright\scriptsize
  1. Typed ``icc world cup'' into search bar \\
  2. Clicked ``STANDINGS'' tab in ICC T20 World Cup widget \\
  3. Scrolled through standings table \\
  4. Clicked ``IND'' (India) in standings \\
  5. Switched to W\&B dashboard tab \\
  6. Scrolled through W\&B charts \\
  7. Clicked ``train/loss'' chart in W\&B \\
  8. Clicked chat with ``Omar Shaikh'' in Slack
  \end{minipage} \\
  \midrule
  % --- 0.98: near-perfect, only minor paraphrasing ---
  0.98 &
  \begin{minipage}[t]{\linewidth}\raggedright\scriptsize
  1. Clicked ``rouge-score'' on PyPI, copied install command \\
  2. Ran \texttt{uv pip install rouge-score} in terminal \\
  3. Scrolled through ``rouge-score'' PyPI page \\
  4. Ran \texttt{accelerate launch} training command \\
  5. Scrolled through ``rouge-score'' PyPI page \\
  6. Switched between W\&B, Google Doc, and terminal \\
  7. Typed training command, cancelled with Ctrl+C \\
  8. Ran \texttt{cd sim-v2} and \texttt{git pull}
  \end{minipage}
  &
  \begin{minipage}[t]{\linewidth}\raggedright\scriptsize
  1. Clicked ``rouge-score'' on PyPI, copied install command \\
  2. Ran \texttt{uv pip install rouge-score} in terminal \\
  3. Scrolled through ``rouge-score'' PyPI page \\
  4. Ran \texttt{accelerate launch} with dataset arguments \\
  5. Scrolled through ``rouge-score'' PyPI page \\
  6. Switched between W\&B, Google Doc, and terminal \\
  7. Typed CUDA training command, cancelled with Ctrl+C \\
  8. Ran \texttt{cd sim-v2} and \texttt{git pull}
  \end{minipage} \\
  \bottomrule
  \end{tabularx}
  \end{table*}

\end{document}